\pdfoutput=1
\documentclass[11pt]{article}
\usepackage[final]{acl}

\usepackage{subcaption}
\usepackage{times}
\usepackage{latexsym}
\usepackage[T1]{fontenc}
\usepackage[utf8]{inputenc}
\usepackage[normalem]{ulem} 
\usepackage{microtype}
\usepackage{inconsolata}
\usepackage{textcomp}
\usepackage{microtype}
\usepackage{graphicx}
\usepackage{import}
\usepackage{layout}
\usepackage{tabularx, makecell}
\newcolumntype{Y}{>{\centering\arraybackslash}X}
\usepackage{booktabs}
\usepackage{ragged2e} 
\usepackage{mathrsfs}
\usepackage{amssymb} 
\usepackage{url}
\usepackage{xspace,paralist}
\usepackage{times,latexsym}
\usepackage{amsmath}
\usepackage{appendix}
\usepackage{comment} 
\usepackage{enumitem}
\usepackage{makecell}
\usepackage{multirow}
\usepackage{xcolor}
\usepackage{cleveref}
\usepackage{todonotes}
\usepackage{longtable,supertabular}
\usepackage[T2A,T1]{fontenc}
\usepackage[russian,english]{babel}
\usepackage{array}
\usepackage{CJKutf8}
\usepackage{float}
\usepackage{adjustbox}

\usepackage{enumitem} 
\usepackage{fancyvrb}  
\usepackage{fvextra}  
\usepackage{amssymb}
\usepackage{adjustbox}
\newcommand{\gain}[1]{\textcolor{blue}{#1}}
\definecolor{dropcolor}{RGB}{170,30,30}

\usepackage{colortbl}
\usepackage[most]{tcolorbox}

\newcolumntype{H}{>{\setbox0=\hbox\bgroup}c<{\egroup}@{}}

\newcommand{\model}{\texttt{SCOUT}}

\usepackage{titlesec}
\usepackage[T2A,T1]{fontenc}

  \DeclareFontFamilySubstitution{T2A}{\familydefault}{Tempora-TLF}%

\begin{document}

\title{Cascaded Retrieval with Cross-Modal Query Refinement for\\Efficient Visual Document Search}

\author{
  \textbf{Diana Turmakhan}$^\heartsuit$ \quad
  \textbf{Aseem Srivastava}$^\heartsuit$ \quad
  \textbf{Vivek Gupta}$^\diamond$ \quad
  \textbf{Iryna Gurevych}$^{\heartsuit\P}$ \quad
  \textbf{Thamar Solorio}$^\heartsuit$ \\[1ex]
  $^\heartsuit$ Mohamed bin Zayed University of Artificial Intelligence 
  $^\diamond$ Arizona State University \\
  $^\P$ Ubiquitous Knowledge Processing Lab (UKP Lab), Department of Computer Science
}
\maketitle

\begin{abstract}
Retrieving information from visually rich documents requires matching queries against pages that contain complex evidence in tables, figures, and layout-dependent structures. Current vision-language retrievers address this but incur query-time costs that grow with corpus size and are impractical to scale. We propose \model, a cascaded two-track retrieval framework that separates retrieval performance from exhaustive search. The visual track first uses a lightweight parent retriever to filter the corpus to a compact candidate pool, then applies a stronger multi-vector reranker only to those candidates. In parallel, a semantic track retrieves over hierarchical page and document summaries generated offline to provide a signal complementary to visual matching. The outputs are fused through cross-modal query refinement, which uses semantic scores to adjust the visual query representation before final ranking. We evaluate \model\ on 4 major benchmarks against 6 baselines and observe a significant query-time latency improvement up to 80.7$\times$ (28.8$\times$ on average) relative to the best baseline, while exceeding state-of-the-art retrieval performance by up to $7.8\%$ NDCG across all benchmarks.
\end{abstract}

\section{Introduction}

\begin{figure}[t]
    \centering
    \includegraphics[width=\linewidth]{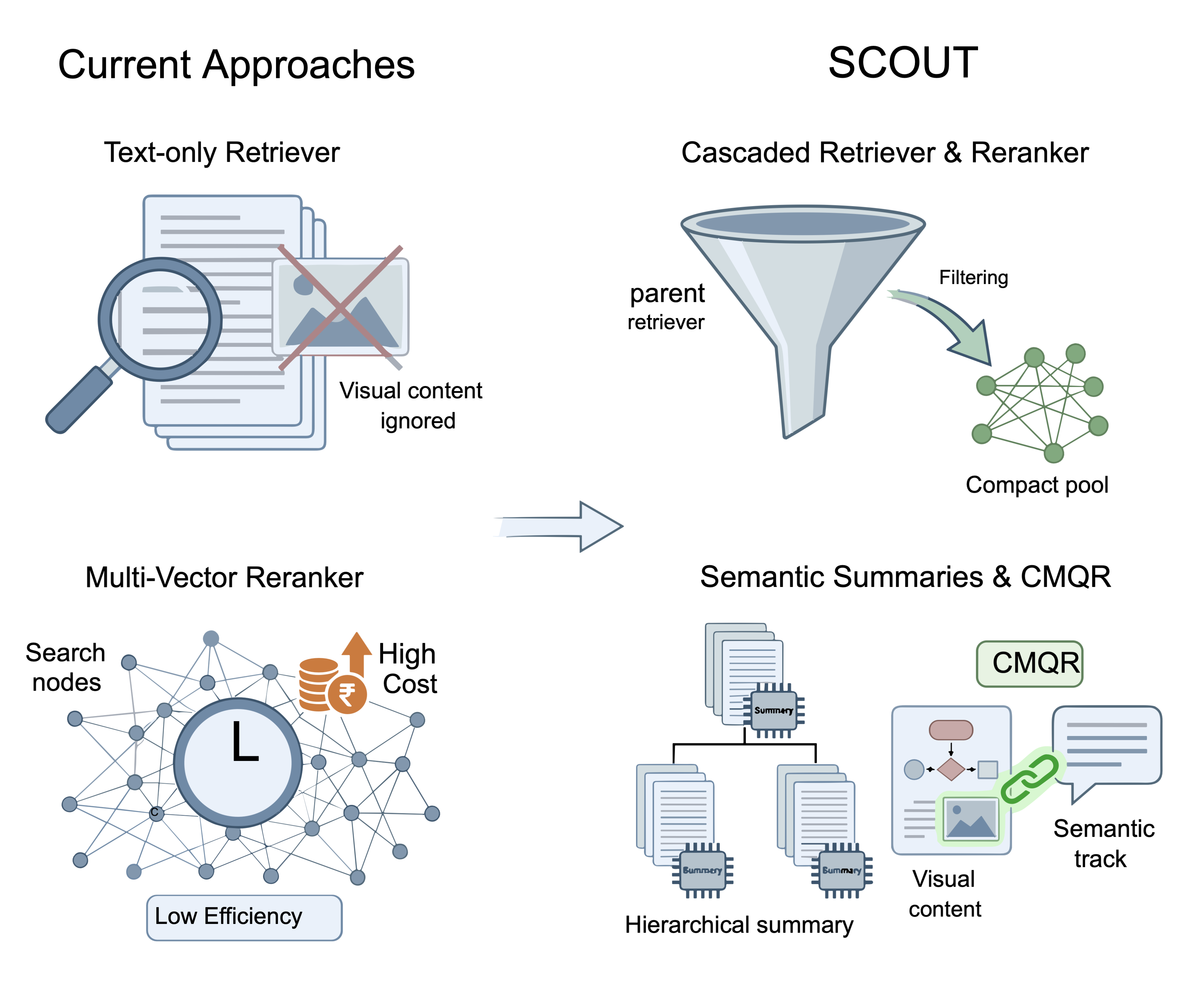}
    \caption{\textbf{Accurate visual retrieval without exhaustive search.} Text-only retrievers ignore visual content, and multi-vector retrievers match every page at high query-time cost. \model{} applies late interaction only to a compact candidate pool and refines the visual query with offline summaries.}
    \label{fig:intro_example}
\end{figure}

Document retrieval is a well-studied problem \cite{abootorabi-etal-2025-ask, Tanaka_2025_CVPR}, yet retrieving visually rich documents remains challenging in open-domain settings that require searching across an entire corpus rather than a predefined collection.
As illustrated in Figure \ref{fig:intro_example}, a query may require evidence distributed across multiple pages of a large report collection. This evidence often appears in tables, figures, or layout-dependent passages, while many pages in the corpus may share similar visual structures. Consequently, even a strong visual retriever that scores every page can fail to place all relevant pages within a practical retrieval window.
This setting highlights two requirements for a practical retriever: (a) matching fine-grained evidence within each page, and (b) doing so at a query-time cost that remains affordable as the corpus grows.

This challenge is relevant to enterprise collections that contain complex information in tables, figures, and layout-dependent structures that text extraction does not preserve completely \cite{huang2022layoutlmv3, cho2025m3docvqa}. Consequently, text-based retrieval can operate on incomplete representations. Our experiments show that retrieving over extracted page text rather than page images reduces NDCG@10 by up to $13.5$ points when using the same embedding backbone (Table \ref{tab:vidoseek_slidevqa_results}). Recent vision-language retrievers address this limitation by embedding document pages directly as images and have achieved strong performance on visual document retrieval benchmarks \cite{ICLR2025_99e9e141, ICLR2025_3640a199, nomic2025colnomic}. 
However, their high query-time latency remains a practical bottleneck for real-world use cases.
In typical enterprise settings, documents are indexed infrequently, while retrieval may occur several times per user request \cite{wang2025vidorag}. 
Retrieval costs therefore accumulate continuously, raising the question of how such systems can scale.
Table \ref{tab:gap} highlights three major gaps in the existing approaches:
(a) \emph{cost:} multi-vector visual retrievers match every query against every page, so query cost grows with the corpus~\cite{NEURIPS2023_31d99727, santhanam2022plaid, 10.1145/3626772.3657769}, while lighter single-vector models are faster but lose fine-grained accuracy \cite{nussbaum2025nomicembedtrainingreproducible};
(b) \emph{recall:} cascades cut cost by filtering before reranking, but a page missed by the filter cannot be recovered later \cite{qu-etal-2021-rocketqa}, which is most likely when many pages look alike; and
(c) \emph{fusion:} textual evidence complements visual matching, but rank-level fusion discards how that evidence relates to the query, and query-level refinement \cite{uzan2026guided} still applies multi-vector scoring to the full corpus.

Our observation is that these gaps can be closed together because the signals are complementary and exhaustive token-level matching is needed only to separate a few visually similar candidates. We therefore propose \model\footnote{{\bf \model}: {\bf S}caled {\bf C}ascade with cr{\bf O}ss-modal q{\bf U}ery refinemen{\bf T}}, a cascaded two-track multimodal retrieval framework. {\em Track A} is a visual cascade: a lightweight single-vector retriever filters the corpus to a small pool of $K$ pages, which a stronger multi-vector model then reranks with late interaction. {\em Track B} retrieves over hierarchical document- and page-level summaries generated once offline, so pages missed by Track A can still enter the pool. The two tracks are combined through cross-modal query refinement (CMQR) \cite{uzan2026guided}, which uses the semantic scores to adjust the visual query representation at test time before the final ranking. 
The costly late-interaction and refinement phases are independent of corpus size and work on a fixed candidate pool.
We evaluate \model\ on four benchmarks, ViDoRe (eight domain-specific subsets), ViDoSeek, SlideVQA, and REAL-MM-RAG (four domain-specific subsets), against six baseline methods.
\model\ matches or exceeds the strongest exhaustive retriever in retrieval quality (+3.1\% average NDCG@10) while improving query latency by 28.8$\times$ on average (up to 80.7$\times$). These gains carry over to downstream answer generation and hold under query rephrasing.
We summarize our contributions as follows:
\begin{itemize}[leftmargin=*, noitemsep]
    \item We propose \model, a cascaded two-track multimodal retrieval framework that reduces query-time latency 
    up to $80.7×$ ($28.8×$ on average) over exhaustive multi-vector search.

    \item \model\ outperforms retrieval performance across all 4 benchmarks, with gains of up to $7.8\%$ NDCG@5 in a fraction of the query-time cost.

    \item We present a thorough analysis of our findings through ablations, candidate-pool sensitivity, robustness to perturbation, and a case study.
\end{itemize}

\begin{table}[t]
\centering
\small
\setlength{\tabcolsep}{3.2pt}
\resizebox{\columnwidth}{!}{
\begin{tabular}{lcccc}
\toprule
\textbf{Approach} & \textbf{Vis.} & \textbf{Txt.} & \textbf{Fus.} & \textbf{Cost} \\
\midrule
\rowcolor{gray!15}\multicolumn{5}{l}{\textit{Single signal}} \\
\hspace{0.8em}Text dense (OCR / summ.)   & $\times$ & \checkmark & --    & --      \\
\hspace{0.8em}Single-vector visual         & page     & $\times$   & --    & --      \\
\hspace{0.8em}Multi-vector visual          & token    & $\times$   & --    & $O(N)$  \\
\midrule
\rowcolor{gray!15}\multicolumn{5}{l}{\textit{Efficiency-oriented}} \\
\hspace{0.8em}Efficient multi-vector index & token    & $\times$   & --    & $O(N')$ \\
\hspace{0.8em}Text cascade                 & $\times$ & \checkmark & --    & $O(K)$  \\
\midrule
\rowcolor{gray!15}\multicolumn{5}{l}{\textit{Multimodal}} \\
\hspace{0.8em}Rank-level hybrid (RRF)      & token    & \checkmark & rank  & $O(N)$  \\
\hspace{0.8em}GQR                          & token    & \checkmark & query & $O(N)$  \\
\rowcolor{yellow!34}\hspace{0.8em}\textbf{\model\ (ours)} & \textbf{token} & \textbf{\checkmark} & \textbf{query} & $\mathbf{O(K)}$ \\
\bottomrule
\end{tabular}
}
\caption{{\bf Taxonomy of prior approaches.} \textbf{Vis.}/\textbf{Txt.}: visual/textual evidence used (page: one vector per page; token: token- or patch-level). \textbf{Fus.}: how modalities are combined (rank: rank-level; query: query-level refinement). \textbf{Cost}: number of pages on which the costly stage (late interaction or reranking) runs, where $N$ is the corpus size, $N' < N$ the candidates kept by index pruning, and $K \ll N$ the first-stage pool.}
\label{tab:gap}
\end{table}

\section{Related Work}
\paragraph{Textual and Visual Retrieval for Document RAG.}
Visual document retrieval (VDR) aims to retrieve relevant pages from visually rich documents containing dense layouts, figures, tables, and charts. 
Traditional pipelines rely on OCR or extracted text followed by lexical or dense retrieval~\citep{karpukhin2020dense,tanaka2021visualmrc}; recent dense models such as Linq-Embed, Qwen3-Embedding, and Jina-Text provide strong OCR- or summary-based text baselines~\citep{choi2024linq,zhang2025qwen3,gunther-etal-2025-jina}. 
However, text-only retrieval can miss evidence encoded in layout and visual structure~\cite{huang2022layoutlmv3}. 
Native visual and multimodal retrievers such as Jina Vision and ColNomic-7B embed page images directly, while late-interaction models such as ColEmbed-3B-v2 use MaxSim-style token- or patch-level scoring~\citep{gunther-etal-2025-jina,nomic2025colnomic,moreira2026nemotroncolembedv2topperforming}. 
Our pipeline is built in this direction where we rerank a compact set of candidate pages and fuse the results with summary-based feedback, avoiding full-corpus multi-vector scoring. 

\paragraph{Hybrid Fusion and Query Refinement.}
Hybrid retrieval combines complementary retrievers, such as lexical, dense-text, and visual models~\citep{dos2015learning,kuzi2020leveraging}. 
Reciprocal Rank Fusion (RRF) is a simple rank-level method for combining retrieval outputs~\citep{cormack2009reciprocal}, but it does not use the geometry of the embedding spaces. 
Test-time query refinement instead updates the query representation using retrieved evidence, building on pseudo-relevance feedback in dense retrieval~\citep{yu2021improving,sung-etal-2023-optimizing}. 
Guided Query Refinement (GQR) refines visual query embedding using feedback from a complementary text retriever~\cite{uzan2026guided}. 
We apply this idea in a cascaded setting, where a lightweight first stage forms the candidate pool before ColEmbed-style reranking and  Qwen3-guided refinement.

\paragraph{Efficient Cascaded Multi-Vector Retrieval.}
Cascaded retrieval balances efficiency and accuracy by using a fast first-stage retriever followed by a stronger reranker \citep{wang2011cascade,nogueira2019passage}. 
This is especially important for multi-vector retrieval, where late-interaction scoring over the full corpus is expensive. 
Prior work such as PLAID and EMVB improves efficiency within multi-vector indexes through pruning, clustering, or quantization~\citep{santhanam2022plaid,nardini2024efficient}. 

In summary, prior work addresses efficiency, recall, or multimodal evidence in isolation (c.f. Table \ref{tab:gap}). Multi-vector indexes speed up scoring within a single modality, cascades are limited by first-stage recall, and hybrid fusion adds textual evidence but still scores the full corpus. \model\ brings novelty by cascading this pipeline for optimized retrieval. Cheap visual filtering and offline textual summaries select a small pool of $K$ pages, so costly late interaction and query refinement run only on that pool and their cost is independent of corpus size, while the textual evidence both recovers missed pages and shapes the final ranking.

\enlargethispage{\baselineskip}
\section{Methodology}
\label{sec:method}
\begin{figure*}[t]
    \includegraphics[width=1\textwidth]{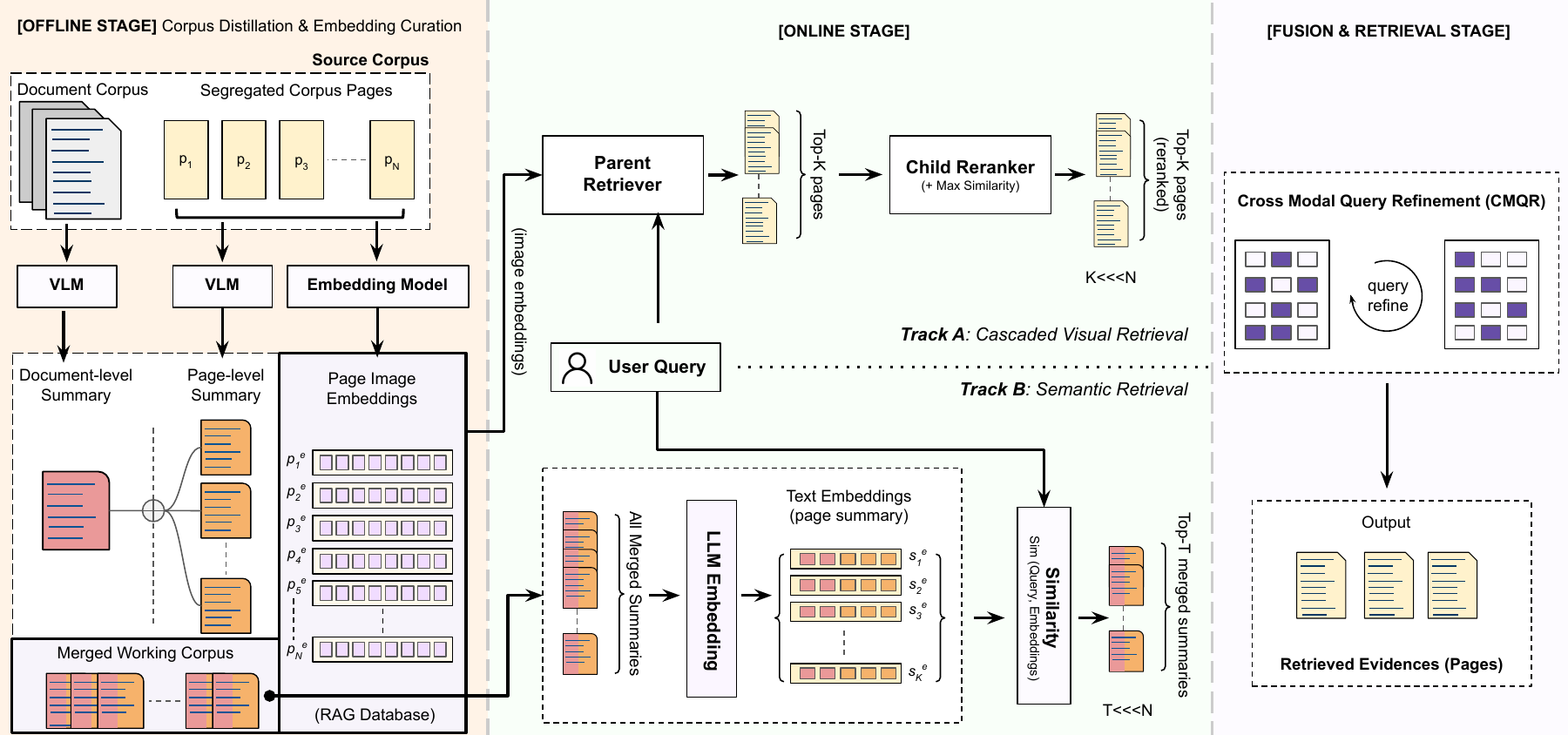}
    \caption{{\bf Architecture of \model.} Left (offline phase): each page in the corpus is dual-encoded as (a) a concatenated doc-level + page-level summary embedding and (b)  a single visual embedding. Center (online phase): a single query q goes through two parallel tracks: Track A performs an efficient single-vector retrieval to top-K candidates, then applies expensive multi-vector reranking + max-similarity on the same top-K pages. Track B retrieves top-T summaries through cosine similarity between summary embeddings and query. Right: Their union enters Cross-Modal Query Refinement (CMQR), where semantic scores form a soft target that refines the visual multi-vector query; the refined query gets re-scored for the final ranking of evidence pages.}
    \label{fig:framework}
    \vspace{-3mm}
\end{figure*}

In this section, we describe \model{}, whose full pipeline is illustrated in Figure~\ref{fig:framework}.

\paragraph{Problem Formulation.}
Let $\mathcal{D} = \{d_1, \ldots, d_N\}$ denote a corpus of $N$ documents. Each document $d_i$ is an ordered sequence of pages $d_i = (p_{i,1}, \ldots, p_{i,m_i})$, where $m_i$ is the number of pages in document $i$. The full set of pages is
\begin{equation}
    \mathcal{P} = \bigcup_{i=1}^{N} \{p_{i,1}, \ldots, p_{i,m_i}\}
    % , 
    % \qquad |\mathcal{P}| = M.
\end{equation}
Each page $p \in \mathcal{P}$ is represented as a rendered image, preserving the original visual layout. Given a query $q$, the retriever produces a ranked list of pages $\pi_q = (p_{\sigma(1)}, p_{\sigma(2)}, \ldots)$ such that pages relevant to $q$ appear at the top. We denote the gold relevance set for query $q$ as $\mathcal{R}_q \subseteq \mathcal{P}$.

\subsection{Proposed Framework: \textsc{\model}}
\label{sec:framework}

\model\ is a cascaded framework for visual document retrieval.
To decouple the cost of corpus processing from query-time computation, we segregate the framework into two phases, online and offline.
% The framework operates in two phases. 
In the offline phase, we construct a compressed representation of the corpus by generating hierarchical textual summaries for each document and page, as described next. In the online phase, a query is processed through two parallel tracks: (a) {\bf Track A} is a cascaded visual track that operates on rendered page images (Section \ref{sec:visual}), and (b) {\bf Track B} is a semantic track that operates on the pre-computed summaries (Section \ref{sec:semantic}). The outputs of both tracks are jointly refined to produce the final ranked list.

\paragraph{Corpus Distillation.}
\label{sec:distillation}
At first, each document $d_i$ and each of its pages $p_{i,j}$ are independently passed to a VLM to generate textual summaries. This step yields two levels of representation: (a) a document-level summary $\delta_i$ to capture the global topic, structure, and key content of document $d_i$; (b) a page-level summary $\phi_{i,j}$ to capture the local content of page $p_{i,j}$, including its tables, figures, and named entities. The working summary text for each page is formed by concatenating the two levels:
\begin{equation}
    \sigma_{i,j} = [\,\delta_i \;\|\; \phi_{i,j}\,],
\end{equation}
The document-level summary $\delta_i$ is prepended to each page summary to provide document context that a page-level description alone may lack, particularly relevant for multi-hop queries. Each summary $\sigma_{i,j}$ is encoded into a dense vector $\mathbf{e}^{\text{sum}}_{i,j} \in \mathbb{R}^{D_s}$ using Qwen3-Embedding \cite{zhang2025qwen3} and indexed for retrieval. In parallel, each page image $p_{i,j}$ is encoded and indexed for the visual track. We present the prompts used for summary generation in Fig. \ref{fig:doc_indexing_prompt} and Fig. \ref{fig:page_indexing_prompt} (c.f. Appendix).

\subsubsection{Cascaded Visual Track (Track A)}
\label{sec:visual}
The visual track retrieves and reranks pages using rendered page images in two stages.

\paragraph{{Stage 1}: Parent Retriever}
Each page $p \in \mathcal{P}$ is encoded offline into a single-vector visual embedding using a lightweight vision-language retriever. At query time, the query $q$ is encoded into $\mathbf{e}^{1B}_q$ and pages are ranked by cosine similarity:
\begin{equation}
    s^{1B}(q, p) = \frac{{\mathbf{e}^{1B}_q}^{\top} \mathbf{e}^{1B}_p}{\|\mathbf{e}^{1B}_q\| \|\mathbf{e}^{1B}_p\|}.
\end{equation}
The top-$K$ pages are retained as the visual candidate set:
\begin{equation}
    \mathcal{C}^{\text{vis}}_q = \operatorname{Top-K}_{p \in \mathcal{P}}\; s^{1B}(q, p).
\end{equation}
Since each page contributes a single vector, scoring over the full corpus reduces to one dense inner product computation per query, making this stage efficient at corpus scale.

\smallskip
\noindent\textbf{Stage 2: Child Reranker.}
The candidate set $\mathcal{C}^{\mathrm{vis}}_q$ is reranked using a state-of-the-art late-interaction multi-vector visual retriever $f_{\mathrm{mv}}$, following the ColBERT paradigm~\citep{khattab2020colbert}. For each page $p$, the encoder produces a sequence of per-token embeddings:
\begin{equation}
    \mathbf{V}_p = [\mathbf{v}^{(1)}_p, \ldots, \mathbf{v}^{(L_p)}_p] \in \mathbb{R}^{L_p \times d}.
\end{equation}
The query is encoded as $\mathbf{Q}_q = [\mathbf{q}^{(1)}, \ldots, \mathbf{q}^{(L_q)}] \in \mathbb{R}^{L_q \times d}$. Relevance is computed via MaxSim:
\begin{equation}
    s^{\text{vis}}(q, p;\, \mathbf{Q}_q)
    =
    \frac{1}{L_q}
    \sum_{i=1}^{L_q}
    \max_{j \in [L_p]}
    \mathbf{q}^{(i)\top} \mathbf{v}^{(j)}_p .
\end{equation}
Since late-interaction scoring is applied only over $\mathcal{C}^{\text{vis}}_q$, the computational cost of this stage is bounded by $O(K)$ regardless of corpus size. The reranked candidate set forms the visual output $\mathcal{O}^{\text{vis}}_q$, with $|\mathcal{O}^{\text{vis}}_q| \le K$.

\subsubsection{Semantic Retrieval (Track B)}
\label{sec:semantic}
We retrieve pages based on their hierarchical summaries $\sigma_{i,j}$,  complementary to the visual track for fine-grained nuances that visual matching may not surface. To address this, the query $q$ is encoded into $\mathbf{e}^{\text{sum}}_q \in \mathbb{R}^{D_s}$ using the same embedding model used for indexing. Each page is scored by cosine similarity against its stored summary embedding.
\begin{equation}
    s^{\text{sum}}(q, p) = \frac{{\mathbf{e}^{\text{sum}}_q}^{\top} \mathbf{e}^{\text{sum}}_p}{\|\mathbf{e}^{\text{sum}}_q\| \|\mathbf{e}^{\text{sum}}_p\|}.
\end{equation}
The top-$T$ pages by summary score act as the semantic output $\mathcal{O}^{\text{sum}}_q$.

\subsubsection{Cross-Modal Query Refinement.}
\label{sec:gqr}
\model's two parallel tracks yield complementary rankings: the visual track captures layout, figures, and spatial structure; the semantic track captures textual and numerical content surfaced through the summaries. Traditionally, combining these at the rank level discards information about how semantic scores relate to individual visual query token representations.
Hence, we apply a novel representation-level fusion using Cross Modal Query Refinement (CMQR) adapted from Guided Query Refinement \cite{uzan2026guided}, which uses the semantic track's scores to refine the visual query embedding before final scoring.

Let $\mathcal{U}_q = \mathcal{O}^{\text{vis}}_q \cup \mathcal{O}^{\text{sum}}_q$ be the union of the two tracks' outputs. Starting from the initial multi-vector query embedding $\mathbf{Q}^{(0)}_q$ produced by the visual encoder, the semantic scores over $\mathcal{U}_q$ are converted into a soft target distribution:
\begin{equation}
    \mathbf{t}_q = (1-\alpha)\cdot\rho_\tau\!\left(\mathbf{s}^{\text{vis},0}\right)
                 + \alpha \cdot\rho_\tau\!\left(\mathbf{s}^{\text{sum}}\right)
\end{equation}
where, $\rho_\tau(\mathbf{x}) = \operatorname{softmax}(\mathbf{x}/\tau)$ denote the temperature-scaled softmax and $\mathbf{s}^{\text{vis},0}$ and $\mathbf{s}^{\text{sum}}$ are the visual and semantic score
vectors over $\mathcal{U}_q$, $\tau$ controls distribution sharpness, and $\alpha$ balances
the two signals. The query embedding is optimized so that the resulting visual score
distribution agrees with $\mathbf{t}_q$, yielding a refined embedding $\mathbf{Q}^{\star}_q$.
The final ranked list is produced by rescoring $\mathcal{U}_q$ with the refined query:
\begin{equation}
    \pi^{\text{final}}_q = \operatorname{rank}_{p \in \mathcal{U}_q}\; s^{\text{vis}}(q, p;\, \mathbf{Q}^{\star}_q).
\end{equation}
The optimization operates only over $\mathcal{U}_q$, which contains at most $K+T$ pages, keeping the cost of this stage independent of corpus size.
Each page in $\pi^{\text{final}}_q$ carries an index $(i,j)$ from the offline phase, which maps directly to both the structured summary $\sigma_{i,j}$ and the original page image $p_{i,j}$, allowing either to be returned as evidence for downstream tasks.

\section{Experiments}
\label{sec:experiments}
In this section, we summarize benchmark, baseline, and metric selection.
\paragraph{Datasets.}
We evaluate on ViDoRe v3 \cite{loison2026vidore}, a domain-diverse visual document retrieval benchmark spanning multiple document types and application areas. To test robustness across different document collections, we additionally evaluate on ViDoSeek \cite{wang2025vidorag}, refined SlideVQA \cite{tanaka2023slidevqa}, and REAL-MM-RAG \cite{wasserman-etal-2025-real}. These datasets cover visually rich pages with tables, figures, charts, slides, and dense layouts, making them suitable for comparing text-only, visual, and multi-vector retrieval methods. 

\paragraph{Evaluation.} We evaluate efficiency using average query latency (ms/query), measured over online retrieval, reranking, fusion, and refinement after offline indexing. Retrieval quality is measured using NDCG@5 and NDCG@10.

\paragraph{Baselines.}
We compare against dense text, visual, and multi-vector retrieval baselines. 
For text-only retrieval, we evaluate Linq-Embed-Mistral~\cite{choi2024linq}, Qwen3-Embedding \cite{zhang2025qwen3}, and Jina-v4 text \cite{gunther-etal-2025-jina} embeddings using OCR-derived text. For visual retrieval, we evaluate models that operate on page images, including Jina-v4 vision, ColNomic-7B~\cite{nomic2025colnomic} and  ColEmbed-3B-v2~\cite{DBLP:journals/corr/abs-2507-05513}. Table \ref{tab:combined_model_methodology} in summarizes all methods.

\section{Results}
\label{sec:results}
In this section, we report retrieval performance, analyze the contribution of each module in \model, and examine the resulting accuracy--latency trade-offs. Additional details are provided in Appendix~\ref{app:additional_results}.

\subsection{Performance Comparison}
\label{sec:results:perf}

The performance of \model\ across four benchmarks is summarized in Table \ref{tab:vidoseek_slidevqa_results}. \model\ outperforms (or levels up) SOTA performance across all benchmarks while retrieving at a fraction of its query-time cost. SOTA (in our case, ColEmbed-3B-v2) suffers with latencies between $4.5$--$32.8$ sec/query depending on corpus size. \model\ limits late-interaction scoring to Top-K candidate pages, reducing latency to $387$-$553$ milliseconds/query across all benchmarks. This is a gain of $8.4\times$ on \texttt{ViDoRe-HR} to $80.7\times$ on \texttt{SlideVQA}, with an average of $28.8\times$ across all benchmarks  (c.f. Table \ref{tab:latency_detailed}). The Pareto boundary in Figure \ref{fig:latency-plots} (a) and the latency breakdown in Figure \ref{fig:latency-plots} (b) illustrate this efficiency advantage. Also, we report the per-dataset index footprint comparison in Table \ref{tab:index_footprint_corpora}.

Overall, \model\ demonstrates consistent improvement across all benchmarks. 
On \texttt{ViDoRe 3}, \model\ achieves NDCG@5 of $58.85$ and NDCG@10 of $61.31$, compared to $56.65$ and $58.99$ for SOTA, an average improvement of $+3.9$\% respectively. On \texttt{ViDoSeek}, NDCG@5 improves from $85.48$ to $87.41$ ($\Delta{=}+2.3$\%) when compared with the best performer (Jina-vis). On \texttt{SlideVQA}, where the strongest baseline is already near saturation, the gain is modest, but SCOUT still provides a substantial latency reduction (see Table~\ref{tab:vidoseek_slidevqa_results}). To evaluate \model’s performance in downstream tasks, we report performance on the Vidore V3 benchmark in Table \ref{tab:vidore3_accuracy} (c.f. Appendix \ref{sec:downstream}).

\paragraph{Generalization Across Domains and Languages.} Beyond the primary benchmarks, \model\ demonstrates strong generalization across multilingual and domain-specific retrieval settings. Table \ref{tab:vidore_results} summarizes findings on ViDoRe benchmark, where it \model\ achieves the best performance across all English and French domain subsets, with improvements of up to $5.7$\% over the strongest baseline. Whereas on REAL-MM-RAG benchmark, as shown in Table \ref{tab:real_mm_rag_results}, \model\ achieves the best results on three of four subsets, with gains of up to $7.8$\%. Detailed results are provided in Appendix \ref{app:domainLanguage}.

\begin{table}[t]
\centering
\setlength{\tabcolsep}{3.2pt}
\resizebox{\columnwidth}{!}{
\begin{tabular}{lcccccccc}
\toprule
\multirow{2}{*}{\textbf{Model}} & \multicolumn{2}{c}{\textbf{ViDoSeek}} & \multicolumn{2}{c}{\textbf{SlideVQA}} & \multicolumn{2}{c}{\textbf{ViDoRe 3}} & \multicolumn{2}{c}{\textbf{R-MM-RAG}} \\
\cmidrule(lr){2-3} \cmidrule(lr){4-5} \cmidrule(lr){6-7} \cmidrule(lr){8-9}
 & @5 & @10 & @5 & @10 & @5 & @10 & @5 & @10 \\ \midrule
Qwen3-8B & 74.84 & 77.10 & 89.01 & 90.24 & 48.18 & 50.63 & 57.61 & 60.35 \\
Jina-v4 (t) & 75.24 & 77.09 & 91.53 & 92.32 & 51.30 & 53.94 & 52.39 & 55.41 \\
Linq-Emb. & 78.20 & 80.15 & 91.12 & 92.05 & 46.77 & 49.44 & 57.36 & 60.09 \\
Jina-v4 (v) & \underline{85.48} & \underline{86.46} & 93.75 & 94.34 & 53.01 & 55.66 & 66.11 & 68.87 \\
ColNomic & 83.48 & 85.03 & 94.06 & 94.57 & 55.44 & 58.02 & 70.60 & 72.60 \\
ColEmbed & 84.76 & 85.76 & \underline{94.58} & \underline{94.86} & \underline{56.65} & \underline{58.99} & \underline{72.43} & \underline{74.56} \\
\midrule
\rowcolor{yellow!34}\textbf{\model} & \textbf{87.41} & \textbf{88.29} & \textbf{94.72} & \textbf{95.01} & \textbf{58.85} & \textbf{61.31} & \textbf{75.06} & \textbf{76.72} \\
\textit{\footnotesize (Std. error)} & \footnotesize$\pm$0.79 & \footnotesize$\pm$0.71 & \footnotesize$\pm$0.32 & \footnotesize$\pm$0.30 & \footnotesize$\pm$3.71 & \footnotesize$\pm$3.73 & \footnotesize$\pm$4.48 & \footnotesize$\pm$4.14 \\
\midrule
$\Delta(\%)$ & \gain{$\uparrow$ 2.3\%} & \gain{$\uparrow$ 2.1\%} & \gain{$\uparrow$ 0.1\%} & \gain{$\uparrow$ 0.2\%} & \gain{$\uparrow$ 3.9\%} & \gain{$\uparrow$ 3.9\%} & \gain{$\uparrow$ 3.6\%} & \gain{$\uparrow$ 2.9\%} \\
\bottomrule
\end{tabular}
}
\caption{Retrieval performance comparison on ViDoSeek, SlideVQA, ViDoRe 3, and REAL-MM-RAG benchmarks. $\Delta(\%)$ shows the relative improvement wrt best baseline (underlined).}
\label{tab:vidoseek_slidevqa_results}
\vspace{-5mm}
\end{table}

\subsection{Ablation Study}
\label{sec:results:ablation}
We present experiments to show module-wise contribution and selection of candidate pool depth (K).

\paragraph{Module-wise Contribution.}Table \ref{tab:component_ablation} shows the contribution of each module in \model\ evaluated on the average NDCG@10 over ViDoRe 3 dataset. The {\em child reranker} module contributes the most to the performance, removal of which causes the steepest $20.90\%$ drop in NDCG@10. This also attests that while the Parent Retriever efficiently filters the candidate pool, the late-interaction scoring of the child reranker is where the visual track's accuracy is realized. We see consistent behavior for other modules. Without Track A, the performance drops by $12.35\%$ and shows that the visual track carries the primary retrieval signal. Similarly, removing Track B alone results in a $7.01\%$ drop. At last, replacing CMQR with common fusion method, RRF, yields a $6.28\%$ drop and indicates that rank-level fusion discards as much signal as the absence of the semantic track, whereas CMQR preserves it at the representation level. We discuss the inclusion of summary vs traditional OCR text in Section \ref{sec:summaryVSocr}.

\paragraph{Candidate Pool Depth.}
\label{ablation:K_sweep}

\begin{table}[t]
\centering
\resizebox{\columnwidth}{!}{
\begin{tabular}{lr}
\toprule
\textbf{Variant} & \textbf{Avg. NDCG@10 ($\Delta\%$)} \\
\midrule
\rowcolor{yellow!34}\textbf{\model} 
& \textbf{61.31} (reference) \\
\midrule
\hspace{1em}w/o Track A 
& 53.73 (\textcolor{red}{$\downarrow 12.35\%$})\\
\hspace{1em}w/o Track B 
& 57.00 (\textcolor{red}{$\downarrow 7.01\%$}) \\
\hspace{2em}$\hookrightarrow$ w/o Child Reranker 
& 48.49 (\textcolor{red}{$\downarrow 20.90\%$}) \\
\hspace{1em}Replace (Track B): Summary$\rightarrow$ OCR & 59.08 (\textcolor{red}{$\downarrow 3.6\%$}) \\
\hspace{1em}Replace (Fusion): CMQR$\rightarrow$ RRF 
& 57.45 (\textcolor{red}{$\downarrow 6.28\%$}) \\
\bottomrule
\end{tabular}
}
\caption{{\bf Ablation Study.} Top-down ablation on the major \model\ modules: Track A (cascaded visual track containing parent retriever and child reranker) and Track B (semantic retriever) in comparison with \model's performance on \texttt{ViDoRe 3}. Additional benchmark ablation in Appendix (c.f. Table \ref{tab:component_ablation_appendix}).}
\label{tab:component_ablation}
\vspace{-3mm}
\end{table}

\begin{figure}[t]
  \centering
  \includegraphics[width=0.98\linewidth,keepaspectratio]{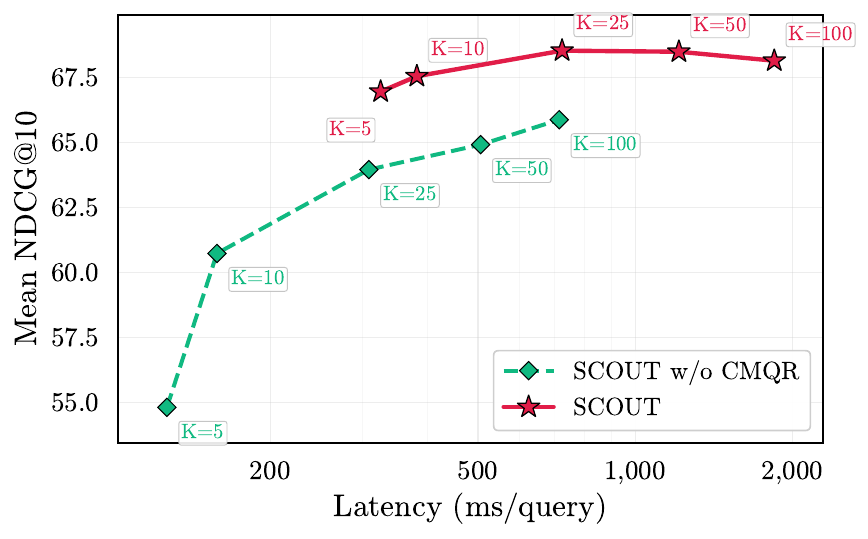}
  \caption{Candidate-depth sweep: Mean NDCG@10 versus latency for different reranker depths $K$.}
  \vspace{-3mm}
  \label{fig:k_sweep}
\end{figure}

% \textcolor{blue}{
Figure \ref{fig:k_sweep} shows mean NDCG@10 against query latency for both \model\ and its variant without CMQR across candidate pool sizes $K \in \{5, 10, 25, 50, 100\}$. Findings show that \model\ reaches near-peak performance at $K{=}10$ and remains stable through $K{=}50$ matching this within $0.04$ points, and slightly drops at $K{=}100$. At the same time, reducing $K$ from  $50$ to $25$ lowers reranking latency from ${\sim}1{,}213$\,ms to ${\sim}724$\,ms, a $40\%$ reduction. The gap between \model\ and the variant without CMQR narrows as $K$ grows, from $12.5$ points at $K{=}5$ to $3.1$ points at $K{=}50$. 
The non-CMQR variant depends more on larger candidate pools to maintain quality, while CMQR achieves stronger rankings from smaller pools by semantically reordering visual candidates. Across all depths, \model\ consistently outperforms the variant without CMQR, indicating that representation-level refinement improves candidate ordering.
All hyperparameters, including the summary depth $T{=}20$, are selected on a held-out validation split; the search spaces and a sensitivity sweep over $T$ are reported in Appendix~\ref{app:hyperparameters}.

\begin{figure}[t]
\centering
\setlength{\tabcolsep}{2px}
\renewcommand{\arraystretch}{1.0}

\begin{tabularx}{\linewidth}{c@{\hspace{4px}}Y}
   \multirow{1}{*}[22pt]{\textit{a)}}
      & \includegraphics[width=0.92\linewidth,keepaspectratio]{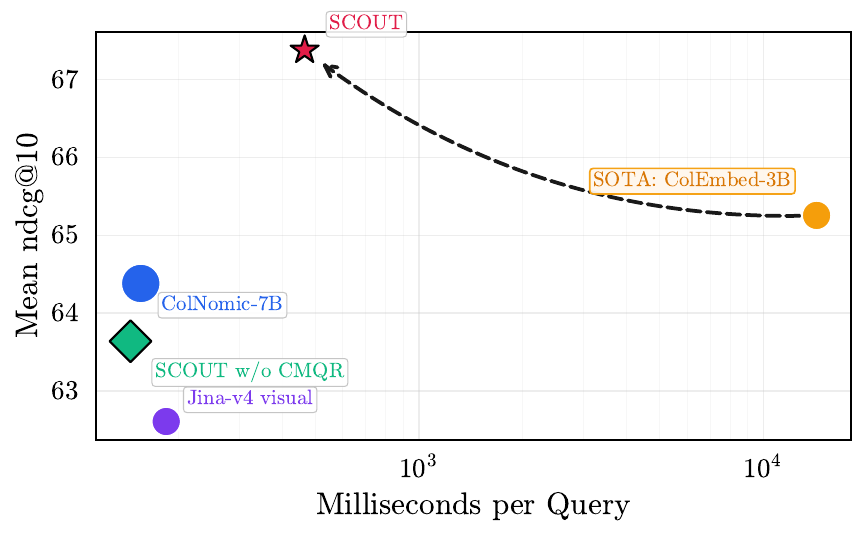} \\
      
   \multirow{1}{*}[22pt]{\textit{b)}}
      & \includegraphics[width=0.92\linewidth,keepaspectratio]{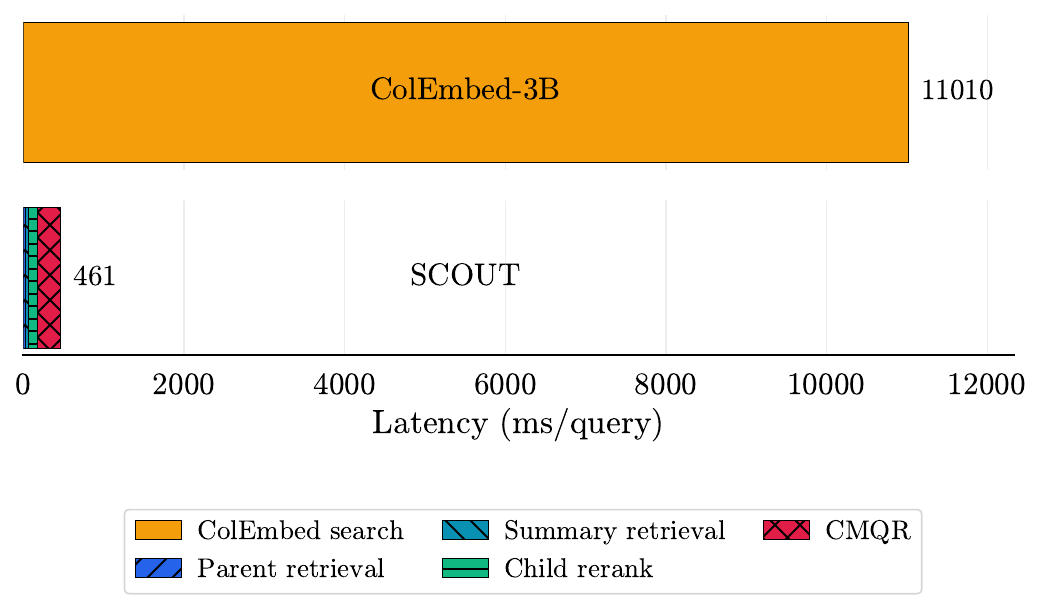} \\
\end{tabularx}
\caption{\textbf{(a)} Latency-quality trade-off for visual retrieval methods. Each point reports macro-averaged NDCG@10 and per-query latency. \textbf{(b)} Per-query latency breakdown comparing exhaustive ColEmbed-3B-v2 retrieval with our cascaded pipeline averaged over the ViDoRe subsets.}
\label{fig:latency-plots}
\vspace{-5mm}
\end{figure}

\section{Analysis}
\label{sec:analysis}
In this section, we analyze the reasoning behind the \model's performance, efficiency, and a case study.

\paragraph{Latency Analysis.}
\label{sec:analysis:latency}
\model\ achieves 468.2 ms/query versus 13, 464.4 ms/query for ColEmbed-3B-v2, a $28.8\times$ efficiency at higher retrieval yield.
The latency breakdown in Figure \ref{fig:latency-plots}(a) shows that the dominant cost is \emph{CMQR refinement} ($\sim$287~ms, $61\%$ of total) over the $K{=}25$ candidate pages, with ColEmbed reranking and Nemotron-Embed-VL-1B~\cite{moreira2026nemotroncolembedv2topperforming} parent retrieval contributing substantially less. Figure \ref{fig:latency-plots-vidoseek-slidevqa} illustrates the latency breakdown on additional benchmarks. CMQR is responsible for per-dataset gains of Finance $+4.40$, HR $+4.58$, Industrial $+5.04$, Finance-FR $+7.78$, Physics $+3.42$, Energy $+4.24$, and CS $+2.48$ NDCG@10 points over the cascade alone (c.f. Table \ref{tab:component_ablation_appendix}). Since both reranking and CMQR operate on at most Top-$K$ pages rather than the full corpus, their cost is independent of corpus size, and the latency improvement is an inherent property of the cascade architecture.

On the other hand, Table \ref{tab:latency_detailed} contextualizes this against the competing retrievers. ColNomic occupies the efficient but lower-quality region at $155.5$ ms/query with NDCG@10$= 64.38$, while ColEmbed defines the slow but strong benchmark at $14,286.4$ ms/query with NDCG@10$= 65.26$. Evidently, Figure \ref{fig:latency-plots}(a) shows that our pipeline sits well above on the Pareto curve at NDCG@10$= 67.37$. 
Further supporting the quality-latency tradeoff under varying $K$ is also analyzed in Section \ref{ablation:K_sweep}, and retrieval index \ footprints are discussed in the Appendix, highlighting active storage reduction between $104$-$337\times$ (c.f. Table \ref{tab:index_footprint_corpora}). We also present the one-time offline summarization cost in Appendix \ref{app:offline_cost}, incurring \$45 for the full ViDoRe corpus and under \$100 across all evaluated corpora (Table~\ref{tab:offline_cost}). We further discuss the practical considerations of such a system in Appendix \ref{sec:pracical_consideration}.

\paragraph{Robustness to Query Perturbation.}
% \label{sec:robustness}
We evaluate robustness to perturbed queries in Table \ref{tab:rephrase_robustness}. \model\ achieves the highest NDCG@10 on the perturbed queries on all four benchmarks and the smallest drop on three of them, losing $4.6$ points on average compared with $6.7$ for ColEmbed and $6.3$ for ColNomic. The gap is largest on REAL-MM-RAG, a benchmark specially built to test robustness to perturbation, where \model\ loses $14.14$ points against $17.25$ and $17.26$. On ViDoSeek it is almost unaffected ($0.07$ points), and on ViDoRe it loses $2.69$ points against $4.64$ and $3.64$. On SlideVQA, queries are straightforward (NDCG@10 near $97$) and perturbation barely removes any vocabulary, so all methods change only marginally. 

\begin{figure}[t]
    \centering
    \includegraphics[width=\linewidth]{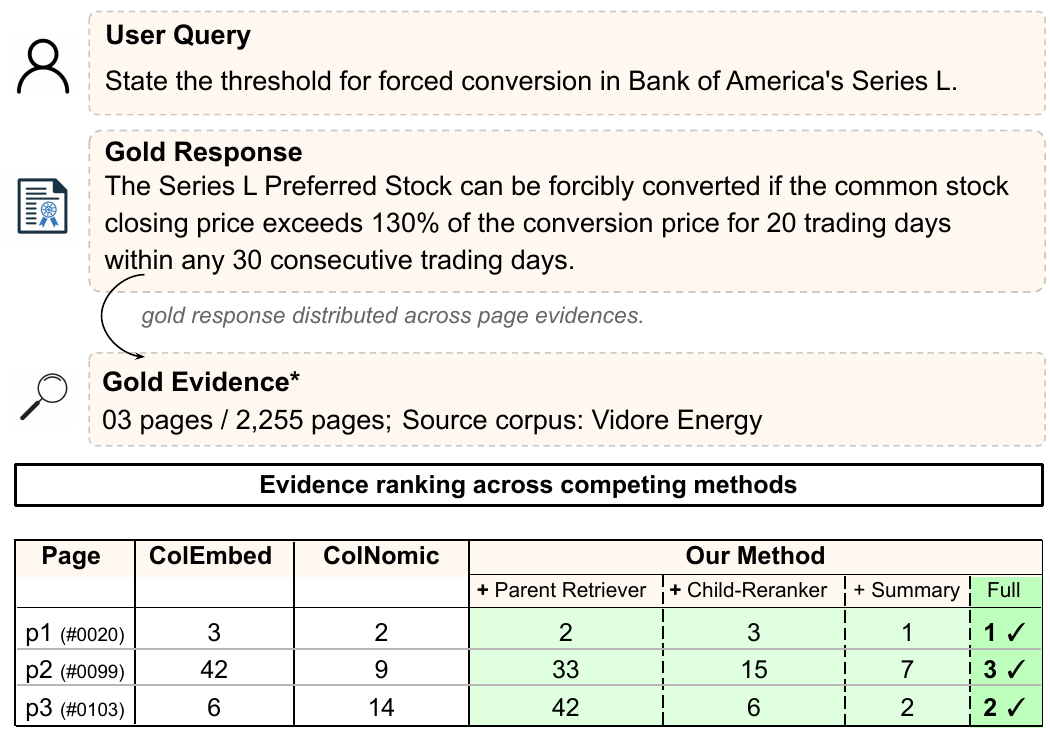}
    \caption{\textbf{Case Study.} A query and its response with evidence distributed across 3 pages out of 2,255 in Vidore-Energy corpus. Rankings are shown at each stage of \model\ against ColEmbed-3B-v2  and ColNomic-7B methods. Our full framework retrieves all three gold pages within  the top-3 ranks ($\checkmark$).} 
    \label{fig:casestudy}
\end{figure}

\paragraph{Case Study.}
\label{analysis:casestudy}
Figure~\ref{fig:casestudy} presents a query from the \texttt{ViDoRe-Energy} corpus  whose answer spans three pages (pg1, pg2, pg3) out of 2,255, making it a case where  no single retrieved page suffices as evidence. The retriever has to fetch all three evidence pages at high ranks, the answer requires numerical precision that visual matching alone may not resolve, and the relevant pages constitute only $0.13\%$ of the corpus. ColEmbed-3B-v2 ranks pg1 at 3rd, pg2 at 42nd, and pg3 at 6th, which recovers pg1 and pg3 reasonably but places pg2 outside any practical retrieval window. ColNomic-7B improves on pg2 (rank 9th) but degrades on pg3 (rank 14th) and fails to consistently surface all three.

For \model, the parent retriever alone ranks pg2 and pg3 poorly (33rd and 42nd) but yields a candidate pool containing all three pages. The child reranker improves pg2 and pg3 substantially (15th and 6th; pg1 moves from 2nd to 3rd), showing that multi-vector late interaction over the pool captures fine-grained visual and textual nuances that single-vector scoring misses. Adding the summary track (Track B) raises pg2 to 7th and pg3 to 2nd, showing that hierarchical summaries supply evidence the visual track alone misses. The full pipeline with CMQR ranks pg1, pg2, and pg3 at 1st, 3rd, and 2nd, placing all three gold pages in the top 3.

\begin{table}[t]
\centering
\setlength{\tabcolsep}{3.2pt}
\resizebox{\columnwidth}{!}{
\begin{tabular}{lcccc}
\toprule
\textbf{Model} & \textbf{ViDoSeek} & \textbf{SlideVQA} & \textbf{ViDoRe 3} & \textbf{R-MM-RAG} \\ \midrule
ColNomic & 88.73 {\footnotesize\textcolor{red}{($\downarrow$3.34)}} & 95.22 {\footnotesize\textcolor{red}{($\downarrow$0.75)}} & 57.41 {\footnotesize\textcolor{red}{($\downarrow$3.64)}} & 72.60 {\footnotesize\textcolor{red}{($\downarrow$17.26)}} \\
ColEmbed & \underline{89.23} {\footnotesize\textcolor{red}{($\downarrow$3.13)}} & \underline{95.29} {\footnotesize\textcolor{red}{($\downarrow$1.67)}} & \underline{58.14} {\footnotesize\textcolor{red}{($\downarrow$4.64)}} & \underline{74.56} {\footnotesize\textcolor{red}{($\downarrow$17.25)}} \\
\midrule
\rowcolor{yellow!34}\textbf{\model} & \textbf{91.00} {\footnotesize\textcolor{red}{($\downarrow$0.07)}} & \textbf{95.52} {\footnotesize\textcolor{red}{($\downarrow$1.58)}} & \textbf{59.94} {\footnotesize\textcolor{red}{($\downarrow$2.69)}} & \textbf{76.72} {\footnotesize\textcolor{red}{($\downarrow$14.14)}} \\
\midrule
\end{tabular}
}
\caption{{\bf Robustness to query perturbation.} The table includes NDCG@10 points on the perturbed queries and the shift from the original scores. Values closer to zero indicate greater robustness. Best scores are in bold and second best are underlined.}
\label{tab:rephrase_robustness}
% \vspace{-5mm}
\end{table}

\paragraph{RRF vs.\ CMQR.}
\label{sec:analysis:gqr_vs_rrf}
We compare two fusion strategies over the same candidate union $\mathcal{U}_q$: reciprocal rank fusion (RRF) and our CMQR (built on GQR). CMQR adds a modest refinement cost but improves the final ranking on every ViDoRe~3 subset and on all four benchmarks (Table~\ref{tab:component_ablation_appendix}; further results in Appendix~\ref{app:rrf_details}). The gap grows as the two tracks' top-10 rankings diverge (Figure~\ref{fig:cmqr_rrf_quintiles}): CMQR's gain over ColEmbed rises from $+1.9\%$ to $+5.7\%$, whereas RRF falls from $-1.2\%$ to $-8.9\%$. The same pattern holds across subsets: CMQR's gain over RRF is largest where the rankings diverge most (Finance-FR $+17.37\%$, Industrial $+12.86\%$, REAL FinReport $+14.49\%$; Spearman $\rho=0.79$ between gain and divergence). We attribute this to when the two signals interact. RRF merges the rankings only after each track has ranked pages independently, so semantic evidence never influences visual matching. CMQR instead uses the semantic scores as soft supervision to refine the visual query token representations directly, so textual evidence guides visual matching before final scoring.

\paragraph{Summary vs Earlier OCR Approaches.}
\label{sec:summaryVSocr}
Replacing the VLM-generated hierarchical summaries with raw page text obtained via OCR \cite{most2025lost, huang2022layoutlmv3} reduces average NDCG@10 by $-3.6$\% while latency remains comparable at ${\sim}547$\,ms. This gap isolates the contribution of the offline distillation step. Raw OCR text preserves some of this information but loses structure and injects noise from layout artifacts and extraction errors. 

\section{Conclusion}
Exhaustive multi-vector search over large document corpora performs well, but is impractical at scale; \model\ addresses this with a cascaded two parallel track framework that combines bounded late-interaction visual retrieval with offline hierarchical summarization and cross-modal query refinement. Across 4 benchmarks, \model\ reduces query-time latency up to $80.7\times$ ($28.8\times$ on average) while exceeding state-of-the-art accuracy by up to $7.8\%$ NDCG. 
Our findings show that the child reranker lifts retrieval performance by $20.9\%$ over the parent retriever alone, and CMQR adds up to $7.78$ NDCG@10 points over the visual cascade by refining the query representation using semantic scores. 
We close with practical considerations showing that \model\ is robust to component choice with interchangeable retrievers.

\clearpage
\newpage
\section{Limitations and Ethical Considerations}
\paragraph{Limitations.}
\model{} can only rank pages that enter its candidate pool. Track~B recovers many pages missed by the parent retriever, but pages ranked low by both tracks remain unrecoverable. The semantic track relies on offline summaries from a proprietary VLM (Gemini~2.5 Flash), which may omit or misrepresent page content. Additionally, the semantic track depends on offline VLM-generated summaries, which may occasionally omit or misrepresent page 
content; higher-quality summarization models can be substituted as they become 
available, given the modular design of \model. Offline indexing is a one-time 
cost excluded from query-time latency and amortizes over all queries, though it may need to be refreshed for frequently updated corpora (which is a lesser frequency practice in enterprise).

\paragraph{Ethical Considerations.}
\model\ is a retrieval framework that does not generate content or model artifacts. Visual document collections may contain sensitive enterprise, financial, legal, or personal information; practitioners should ensure appropriate access control, data retention policies, and privacy-preserving indexing when deploying \model\ in such settings as per the organization policies. As with any retrieval system, retrieved pages should be treated as evidence candidates rather than guaranteed ground truth, and human oversight is encouraged in high-stakes decision-making contexts. Future work should explore privacy-preserving indexing strategies and end-to-end evaluation under realistic user-facing RAG workflows.

\bibliographystyle{acl_natbib}
\bibliography{ref_final}

@inproceedings{wasserman-etal-2025-real,
    title = "{REAL}-{MM}-{RAG}: A Real-World Multi-Modal Retrieval Benchmark",
    author = "Wasserman, Navve and Pony, Roi and Naparstek, Oshri and Goldfarb, Adi Raz and Schwartz, Eli and Barzelay, Udi and Karlinsky, Leonid",
    booktitle = "Proceedings of the 63rd Annual Meeting of the Association for Computational Linguistics (Volume 1: Long Papers)",
    month = jul,
    year = "2025",
    address = "Vienna, Austria",
    publisher = "Association for Computational Linguistics",
    url = "https://aclanthology.org/2025.acl-long.1528/",
    doi = "10.18653/v1/2025.acl-long.1528",
    pages = "31660--31683"
}

@inproceedings{tanaka2021visualmrc,
  title={Visualmrc: Machine reading comprehension on document images},
  author={Tanaka, Ryota and Nishida, Kyosuke and Yoshida, Sen},
  booktitle={Proceedings of the AAAI Conference on Artificial Intelligence},
  volume={35},
  number={15},
  pages={13878--13888},
  year={2021}
}

@inproceedings{dos2015learning,
  title={Learning hybrid representations to retrieve semantically equivalent questions},
  author={Dos Santos, Cicero and Barbosa, Luciano and Bogdanova, Dasha and Zadrozny, Bianca},
  booktitle={Proceedings of the 53rd Annual Meeting of the Association for Computational Linguistics and the 7th International Joint Conference on Natural Language Processing (Volume 2: Short Papers)},
  pages={694--699},
  year={2015}
}

@article{kuzi2020leveraging,
  title={Leveraging semantic and lexical matching to improve the recall of document retrieval systems: A hybrid approach},
  author={Kuzi, Saar and Zhang, Mingyang and Li, Cheng and Bendersky, Michael and Najork, Marc},
  journal={arXiv preprint arXiv:2010.01195},
  year={2020}
}

@inproceedings{cormack2009reciprocal,
  title={Reciprocal rank fusion outperforms condorcet and individual rank learning methods},
  author={Cormack, Gordon V and Clarke, Charles LA and Buettcher, Stefan},
  booktitle={Proceedings of the 32nd international ACM SIGIR conference on Research and development in information retrieval},
  pages={758--759},
  year={2009}
}

@inproceedings{yu2021improving,
  title={Improving query representations for dense retrieval with pseudo relevance feedback},
  author={Yu, HongChien and Xiong, Chenyan and Callan, Jamie},
  booktitle={Proceedings of the 30th ACM International Conference on Information \& Knowledge Management},
  pages={3592--3596},
  year={2021}
}

@inproceedings{sung-etal-2023-optimizing,
    title = "Optimizing Test-Time Query Representations for Dense Retrieval",
    author = "Sung, Mujeen  and
      Park, Jungsoo  and
      Kang, Jaewoo  and
      Chen, Danqi  and
      Lee, Jinhyuk",
    editor = "Rogers, Anna  and
      Boyd-Graber, Jordan  and
      Okazaki, Naoaki",
    booktitle = "Findings of the Association for Computational Linguistics: ACL 2023",
    month = jul,
    year = "2023",
    address = "Toronto, Canada",
    publisher = "Association for Computational Linguistics",
    url = "https://aclanthology.org/2023.findings-acl.354/",
    doi = "10.18653/v1/2023.findings-acl.354",
    pages = "5731--5746"
}

@inproceedings{wang2011cascade,
  title={A cascade ranking model for efficient ranked retrieval},
  author={Wang, Lidan and Lin, Jimmy and Metzler, Donald},
  booktitle={Proceedings of the 34th international ACM SIGIR conference on Research and development in Information Retrieval},
  pages={105--114},
  year={2011}
}

@article{nogueira2019passage,
  title={Passage Re-ranking with BERT},
  author={Nogueira, Rodrigo and Cho, Kyunghyun},
  journal={arXiv preprint arXiv:1901.04085},
  year={2019}
}

@inproceedings{santhanam2022plaid,
  title={PLAID: an efficient engine for late interaction retrieval},
  author={Santhanam, Keshav and Khattab, Omar and Potts, Christopher and Zaharia, Matei},
  booktitle={Proceedings of the 31st ACM International Conference on Information \& Knowledge Management},
  pages={1747--1756},
  year={2022}
}

@inproceedings{nardini2024efficient,
  title={Efficient multi-vector dense retrieval with bit vectors},
  author={Nardini, Franco Maria and Rulli, Cosimo and Venturini, Rossano},
  booktitle={European Conference on Information Retrieval},
  pages={3--17},
  year={2024},
  organization={Springer}
}

@article{choi2024linq,
  title={Linq-embed-mistral technical report},
  author={Choi, Chanyeol and Kim, Junseong and Lee, Seolhwa and Kwon, Jihoon and Gu, Sangmo and Kim, Yejin and Cho, Minkyung and Sohn, Jy-yong},
  journal={arXiv preprint arXiv:2412.03223},
  year={2024}
}

@article{zhang2025qwen3,
  title={Qwen3 embedding: Advancing text embedding and reranking through foundation models},
  author={Zhang, Yanzhao and Li, Mingxin and Long, Dingkun and Zhang, Xin and Lin, Huan and Yang, Baosong and Xie, Pengjun and Yang, An and Liu, Dayiheng and Lin, Junyang and others},
  journal={arXiv preprint arXiv:2506.05176},
  year={2025}
}

@inproceedings{gunther-etal-2025-jina,
    title = "jina-embeddings-v4: Universal Embeddings for Multimodal Multilingual Retrieval",
    author = {G{\"u}nther, Michael  and
      Sturua, Saba  and
      Akram, Mohammad Kalim  and
      Mohr, Isabelle  and
      Ungureanu, Andrei  and
      Wang, Bo  and
      Eslami, Sedigheh  and
      Martens, Scott  and
      Werk, Maximilian  and
      Wang, Nan  and
      Xiao, Han},
    editor = "Adelani, David Ifeoluwa  and
      Arnett, Catherine  and
      Ataman, Duygu  and
      Chang, Tyler A.  and
      Gonen, Hila  and
      Raja, Rahul  and
      Schmidt, Fabian  and
      Stap, David  and
      Wang, Jiayi",
    booktitle = "Proceedings of the 5th Workshop on Multilingual Representation Learning (MRL 2025)",
    month = nov,
    year = "2025",
    address = "Suzhuo, China",
    publisher = "Association for Computational Linguistics",
    url = "https://aclanthology.org/2025.mrl-main.36/",
    doi = "10.18653/v1/2025.mrl-main.36",
    pages = "531--550",
    ISBN = "979-8-89176-345-6"
}

@misc{moreira2026nemotroncolembedv2topperforming,
      title={Nemotron ColEmbed V2: Top-Performing Late Interaction Embedding Models for Visual Document Retrieval}, 
      author={Gabriel de Souza P. Moreira and Ronay Ak and Mengyao Xu and Oliver Holworthy and Benedikt Schifferer and Zhiding Yu and Yauhen Babakhin and Radek Osmulski and Jiarui Cai and Ryan Chesler and Bo Liu and Even Oldridge},
      year={2026},
      eprint={2602.03992},
      archivePrefix={arXiv},
      primaryClass={cs.IR},
      url={https://arxiv.org/abs/2602.03992}, 
}

@article{nomic2025colnomic,
  author       = {{Nomic Team}},
  title        = {Nomic Embed Multimodal: Interleaved Text, Image, and Screenshots for Visual Document Retrieval},
  year         = {2025},
  howpublished = {\url{https://nomic.ai/blog/posts/nomic-embed-multimodal}},
  note         = {Accessed: 2026-05-19}
}

@article{loison2026vidore,
  title={ViDoRe V3: A Comprehensive Evaluation of Retrieval Augmented Generation in Complex Real-World Scenarios},
  author={Loison, Ant{\'o}nio and Mac{\'e}, Quentin and Edy, Antoine and Xing, Victor and Balough, Tom and Moreira, Gabriel and Liu, Bo and Faysse, Manuel and Hudelot, C{\'e}line and Viaud, Gautier},
  journal={arXiv preprint arXiv:2601.08620},
  year={2026}
}

@inproceedings{tanaka2023slidevqa,
  title={Slidevqa: A dataset for document visual question answering on multiple images},
  author={Tanaka, Ryota and Nishida, Kyosuke and Nishida, Kosuke and Hasegawa, Taku and Saito, Itsumi and Saito, Kuniko},
  booktitle={Proceedings of the AAAI Conference on Artificial Intelligence},
  volume={37},
  number={11},
  pages={13636--13645},
  year={2023}
}

@inproceedings{khattab2020colbert,
  title={Colbert: Efficient and effective passage search via contextualized late interaction over bert},
  author={Khattab, Omar and Zaharia, Matei},
  booktitle={Proceedings of the 43rd International ACM SIGIR conference on research and development in Information Retrieval},
  pages={39--48},
  year={2020}
}

@inproceedings{most2025lost,
  title={Lost in ocr translation? vision-based approaches to robust document retrieval},
  author={Most, Alexander and Winjum, Joseph and Bhattarai, Manish and Jones, Shawn and Ranasinghe, Nishath Rajiv and Biswas, Ayan and O'Malley, Dan},
  booktitle={Proceedings of the 2025 ACM Symposium on Document Engineering},
  pages={1--10},
  year={2025}
}

@inproceedings{huang2022layoutlmv3,
  title={Layoutlmv3: Pre-training for document ai with unified text and image masking},
  author={Huang, Yupan and Lv, Tengchao and Cui, Lei and Lu, Yutong and Wei, Furu},
  booktitle={Proceedings of the 30th ACM international conference on multimedia},
  pages={4083--4091},
  year={2022}
}

@inproceedings{karpukhin2020dense,
  title={Dense passage retrieval for open-domain question answering},
  author={Karpukhin, Vladimir and Oguz, Barlas and Min, Sewon and Lewis, Patrick and Wu, Ledell and Edunov, Sergey and Chen, Danqi and Yih, Wen-tau},
  booktitle={Proceedings of the 2020 conference on empirical methods in natural language processing (EMNLP)},
  pages={6769--6781},
  year={2020}
}

@inproceedings{cho2025m3docvqa,
  title={M3docvqa: Multi-modal multi-page multi-document understanding},
  author={Cho, Jaemin and Mahata, Debanjan and Irsoy, Ozan and He, Yujie and Bansal, Mohit},
  booktitle={Proceedings of the IEEE/CVF International Conference on Computer Vision},
  pages={6178--6188},
  year={2025}
}

@inproceedings{wang2025vidorag,
  title={Vidorag: Visual document retrieval-augmented generation via dynamic iterative reasoning agents},
  author={Wang, Qiuchen and Ding, Ruixue and Chen, Zehui and Wu, Weiqi and Wang, Shihang and Xie, Pengjun and Zhao, Feng},
  booktitle={Proceedings of the 2025 Conference on Empirical Methods in Natural Language Processing},
  pages={9124--9145},
  year={2025}
}

@article{wang2024mineru,
  title={Mineru: An open-source solution for precise document content extraction},
  author={Wang, Bin and Xu, Chao and Zhao, Xiaomeng and Ouyang, Linke and Wu, Fan and Zhao, Zhiyuan and Xu, Rui and Liu, Kaiwen and Qu, Yuan and Shang, Fukai and others},
  journal={arXiv preprint arXiv:2409.18839},
  year={2024}
}

@misc{nvidiaingest2024,
  author       = {{NVIDIA Ingest Development Team}},
  title        = {{NVIDIA Ingest}: An Accelerated Pipeline for Document Ingestion},
  year         = {2024},
  howpublished = {\url{https://github.com/NVIDIA/NeMo-Retriever}},
  note         = {Accessed: 2026-05-21}
}

@article{comanici2025gemini,
  title={Gemini 2.5: Pushing the frontier with advanced reasoning, multimodality, long context, and next generation agentic capabilities},
  author={Comanici, Gheorghe and Bieber, Eric and Schaekermann, Mike and Pasupat, Ice and Sachdeva, Noveen and Dhillon, Inderjit and Blistein, Marcel and Ram, Ori and Zhang, Dan and Rosen, Evan and others},
  journal={arXiv preprint arXiv:2507.06261},
  year={2025}
}

@inproceedings{ICLR2025_99e9e141,
 author = {Faysse, Manuel and Sibille, Hugues and Wu, Tony and Omrani, Bilel and Viaud, Gautier and HUDELOT, CELINE and Colombo, Pierre},
 booktitle = {International Conference on Learning Representations},
 editor = {Y. Yue and A. Garg and N. Peng and F. Sha and R. Yu},
 pages = {61424--61449},
 title = {ColPali: Efficient Document Retrieval with Vision Language Models},
 url = {https://proceedings.iclr.cc/paper_files/paper/2025/file/99e9e141aafc314f76b0ca3dd66898b3-Paper-Conference.pdf},
 volume = {2025},
 year = {2025}
}

@inproceedings{abootorabi-etal-2025-ask,
    title = "Ask in Any Modality: A Comprehensive Survey on Multimodal Retrieval-Augmented Generation",
    author = "Abootorabi, Mohammad Mahdi  and
      Zobeiri, Amirhosein  and
      Dehghani, Mahdi  and
      Mohammadkhani, Mohammadali  and
      Mohammadi, Bardia  and
      Ghahroodi, Omid  and
      Baghshah, Mahdieh Soleymani  and
      Asgari, Ehsaneddin",
    editor = "Che, Wanxiang  and
      Nabende, Joyce  and
      Shutova, Ekaterina  and
      Pilehvar, Mohammad Taher",
    booktitle = "Findings of the Association for Computational Linguistics: ACL 2025",
    month = jul,
    year = "2025",
    address = "Vienna, Austria",
    publisher = "Association for Computational Linguistics",
    url = "https://aclanthology.org/2025.findings-acl.861/",
    doi = "10.18653/v1/2025.findings-acl.861",
    pages = "16776--16809",
    ISBN = "979-8-89176-256-5"
}

@inproceedings{ICLR2025_3640a199,
 author = {Yu, Shi and Tang, Chaoyue and Xu, Bokai and Cui, Junbo and Ran, Junhao and Yan, Yukun and Liu, Zhenghao and Wang, Shuo and Han, Xu and Liu, Zhiyuan and Sun, Maosong},
 booktitle = {International Conference on Learning Representations},
 editor = {Y. Yue and A. Garg and N. Peng and F. Sha and R. Yu},
 pages = {21074--21098},
 title = {VisRAG: Vision-based Retrieval-augmented Generation on Multi-modality Documents},
 url = {https://proceedings.iclr.cc/paper_files/paper/2025/file/3640a1997a4c9571cea9db2c82e1fc35-Paper-Conference.pdf},
 volume = {2025},
 year = {2025}
}

@inproceedings{qu-etal-2021-rocketqa,
    title = "{R}ocket{QA}: An Optimized Training Approach to Dense Passage Retrieval for Open-Domain Question Answering",
    author = "Qu, Yingqi  and
      Ding, Yuchen  and
      Liu, Jing  and
      Liu, Kai  and
      Ren, Ruiyang  and
      Zhao, Wayne Xin  and
      Dong, Daxiang  and
      Wu, Hua  and
      Wang, Haifeng",
    editor = "Toutanova, Kristina  and
      Rumshisky, Anna  and
      Zettlemoyer, Luke  and
      Hakkani-Tur, Dilek  and
      Beltagy, Iz  and
      Bethard, Steven  and
      Cotterell, Ryan  and
      Chakraborty, Tanmoy  and
      Zhou, Yichao",
    booktitle = "Proceedings of the 2021 Conference of the North American Chapter of the Association for Computational Linguistics: Human Language Technologies",
    month = jun,
    year = "2021",
    address = "Online",
    publisher = "Association for Computational Linguistics",
    url = "https://aclanthology.org/2021.naacl-main.466/",
    doi = "10.18653/v1/2021.naacl-main.466",
    pages = "5835--5847"
}

@inproceedings{10.1145/3626772.3657769,
author = {Bruch, Sebastian and Nardini, Franco Maria and Rulli, Cosimo and Venturini, Rossano},
title = {Efficient Inverted Indexes for Approximate Retrieval over Learned Sparse Representations},
year = {2024},
isbn = {9798400704314},
publisher = {Association for Computing Machinery},
address = {New York, NY, USA},
url = {https://doi.org/10.1145/3626772.3657769},
doi = {10.1145/3626772.3657769},
booktitle = {Proceedings of the 47th International ACM SIGIR Conference on Research and Development in Information Retrieval},
pages = {152–162},
numpages = {11},
location = {Washington DC, USA},
series = {SIGIR '24}
}

@article{DBLP:journals/corr/abs-2507-05513,
  author       = {Mengyao Xu and
                  Gabriel de Souza Pereira Moreira and
                  Ronay Ak and
                  Radek Osmulski and
                  Yauhen Babakhin and
                  Zhiding Yu and
                  Benedikt Schifferer and
                  Even Oldridge},
  title        = {Llama Nemoretriever Colembed: Top-Performing Text-Image Retrieval
                  Model},
  journal      = {CoRR},
  volume       = {abs/2507.05513},
  year         = {2025},
  url          = {https://doi.org/10.48550/arXiv.2507.05513},
  doi          = {10.48550/ARXIV.2507.05513},
  eprinttype   = {arXiv},
  eprint       = {2507.05513},
  bibsource    = {dblp computer science bibliography, https://dblp.org}
}

@inproceedings{
uzan2026guided,
title={Guided Query Refinement: Multimodal Hybrid Retrieval with Test-Time Optimization},
author={Omri Uzan and Asaf Yehudai and roi pony and Eyal Shnarch and Ariel Gera},
booktitle={The Fourteenth International Conference on Learning Representations},
year={2026},
url={https://openreview.net/forum?id=4GRsedu43K}
}

@InProceedings{Tanaka_2025_CVPR,
    author    = {Tanaka, Ryota and Iki, Taichi and Hasegawa, Taku and Nishida, Kyosuke and Saito, Kuniko and Suzuki, Jun},
    title     = {VDocRAG: Retrieval-Augmented Generation over Visually-Rich Documents},
    booktitle = {Proceedings of the IEEE/CVF Conference on Computer Vision and Pattern Recognition (CVPR)},
    month     = {June},
    year      = {2025},
    pages     = {24827-24837}
}

@inproceedings{NEURIPS2023_31d99727,
 author = {Lee, Jinhyuk and Dai, Zhuyun and Duddu, Sai Meher Karthik and Lei, Tao and Naim, Iftekhar and Chang, Ming-Wei and Zhao, Vincent},
 booktitle = {Advances in Neural Information Processing Systems},
 editor = {A. Oh and T. Naumann and A. Globerson and K. Saenko and M. Hardt and S. Levine},
 pages = {15384--15405},
 publisher = {Curran Associates, Inc.},
 title = {Rethinking the Role of Token Retrieval in Multi-Vector Retrieval},
 url = {https://proceedings.neurips.cc/paper_files/paper/2023/file/31d997278ee9069d6721bc194174bb4c-Paper-Conference.pdf},
 volume = {36},
 year = {2023}
}

@misc{nussbaum2025nomicembedtrainingreproducible,
      title={Nomic Embed: Training a Reproducible Long Context Text Embedder}, 
      author={Zach Nussbaum and John X. Morris and Brandon Duderstadt and Andriy Mulyar},
      year={2025},
      eprint={2402.01613},
      archivePrefix={arXiv},
      primaryClass={cs.CL},
      url={https://arxiv.org/abs/2402.01613}, 
}

\onecolumn 
\appendix
\twocolumn
\section*{Appendix}

\begin{table*}[htbp]
\centering
\setlength{\tabcolsep}{2pt}
\resizebox{\textwidth}{!}{
\begin{tabular}{lcccccccccccccccc}
\toprule
\multirow{3}{*}{\textbf{Model}} & \multicolumn{10}{c}{\textbf{English}} & \multicolumn{6}{c}{\textbf{French}} \\ 
\cmidrule(lr){2-11} \cmidrule(lr){12-17}
 & \multicolumn{2}{c}{CS} & \multicolumn{2}{c}{Finance} & \multicolumn{2}{c}{Pharma} & \multicolumn{2}{c}{HR} & \multicolumn{2}{c}{Ind.} & \multicolumn{2}{c}{Finance} & \multicolumn{2}{c}{Energy} & \multicolumn{2}{c}{Physics} \\
\cmidrule(lr){2-3} \cmidrule(lr){4-5} \cmidrule(lr){6-7} \cmidrule(lr){8-9} \cmidrule(lr){10-11} \cmidrule(lr){12-13} \cmidrule(lr){14-15} \cmidrule(lr){16-17}
 & @5 & @10 & @5 & @10 & @5 & @10 & @5 & @10 & @5 & @10 & @5 & @10 & @5 & @10 & @5 & @10 \\ \midrule
Qwen3-8B & 70.43 & 72.45 & 46.52 & 48.85 & 57.41 & 59.50 & 46.31 & 49.37 & 38.31 & 40.19 & 29.08 & 31.38 & 53.36 & 57.07 & 44.06 & 46.24 \\
Jina-v4 (Text) & 66.61 & 69.08 & 54.68 & 57.25 & 54.01 & 55.89 & 53.34 & 56.27 & 41.13 & 43.17 & 40.63 & 44.21 & 59.25 & 62.17 & 40.77 & 43.44 \\
Linq-Embed & 66.39 & 69.21 & 47.88 & 50.85 & 56.52 & 59.20 & 45.79 & 48.76 & 31.96 & 34.12 & 31.93 & 33.97 & 50.81 & 54.33 & 42.91 & 45.11 \\
Jina-v4 (Vis) & 69.63 & 71.91 & 52.85 & 56.03 & 60.32 & 62.04 & 53.83 & 57.00 & 45.84 & 47.70 & 39.70 & 43.39 & 58.11 & 61.10 & 43.81 & 46.10 \\
ColNomic-7B & 74.18 & 76.93 & 54.29 & 56.88 & 61.07 & 62.89 & 56.83 & 59.91 & 47.69 & 49.56 & \underline{42.19} & \underline{45.89} & 61.10 & 63.69 & \underline{46.18} & \underline{48.44} \\
ColEmbed-3B-v2 & \underline{74.88} & \underline{77.22} & \underline{60.81} & \underline{63.40} & \underline{63.79} & \underline{65.47} & \underline{58.61} & \underline{61.32} & \underline{48.46} & \underline{49.85} & 40.31 & 43.38 & \underline{61.37} & \underline{64.17} & 44.93 & 47.10 \\
\midrule
\rowcolor{yellow!34}\textbf{\model} (ours) & \textbf{75.65} & \textbf{78.13} & \textbf{64.27} & \textbf{66.86} & \textbf{64.69} & \textbf{66.99} & \textbf{61.49} & \textbf{64.12} & \textbf{49.87} & \textbf{51.09} & \textbf{44.42} & \textbf{47.58} & \textbf{62.29} & \textbf{65.21} & \textbf{48.09} & \textbf{50.46} \\
\textit{\footnotesize (Std. error)} & \footnotesize$\pm$0.71 & \footnotesize$\pm$0.64 & \footnotesize$\pm$0.79 & \footnotesize$\pm$0.74 & \footnotesize$\pm$0.72 & \footnotesize$\pm$0.67 & \footnotesize$\pm$0.76 & \footnotesize$\pm$0.71 & \footnotesize$\pm$0.99 & \footnotesize$\pm$0.94 & \footnotesize$\pm$0.89 & \footnotesize$\pm$0.84 & \footnotesize$\pm$0.86 & \footnotesize$\pm$0.81 & \footnotesize$\pm$0.83 & \footnotesize$\pm$0.77 \\
\midrule
$\Delta\%$ vs. Best & \gain{$\uparrow$ 1.0\%} & \gain{$\uparrow$ 1.2\%} & \gain{$\uparrow$5.7\%} & \gain{$\uparrow$5.5\%} & \gain{$\uparrow$1.4\%} & \gain{$\uparrow$2.3\%} & \gain{$\uparrow$ 4.9\%} & \gain{$\uparrow$ 4.6\%} & \gain{$\uparrow$ 2.9\%} & \gain{$\uparrow$ 2.5\%} & \gain{$\uparrow$ 5.3\%} & \gain{$\uparrow$ 3.7\%} & \gain{$\uparrow$ 1.5\%} & \gain{$\uparrow$ 1.6\%} & \gain{$\uparrow$ 4.1\%} & \gain{$\uparrow$ 4.2\%} \\
\bottomrule
\end{tabular}
}
\caption{Fine-grained retrieval performance on multilingual ViDoRe subsets across English and French domains. Results are reported using NDCG@\{5,10\}. \model\ achieves the best performance across all domains and shows generalization to multilingual and domain-specific retrieval settings. Consolidated results are in Table \ref{tab:vidoseek_slidevqa_results}.}
\label{tab:vidore_results}
\end{table*}

\begin{table}[htbp]
\centering
\setlength{\tabcolsep}{4pt}
\resizebox{\columnwidth}{!}{
\begin{tabular}{lcccccccc}
\toprule
\multirow{3}{*}{\textbf{Model}} & \multicolumn{4}{c}{\textbf{Finance}} & \multicolumn{4}{c}{\textbf{Technical}} \\
\cmidrule(lr){2-5} \cmidrule(lr){6-9}
 & \multicolumn{2}{c}{FinSlides} & \multicolumn{2}{c}{FinReport} & \multicolumn{2}{c}{TechReport} & \multicolumn{2}{c}{TechSlides} \\
\cmidrule(lr){2-3} \cmidrule(lr){4-5} \cmidrule(lr){6-7} \cmidrule(lr){8-9}
 & @5 & @10 & @5 & @10 & @5 & @10 & @5 & @10 \\ \midrule
Qwen3-8B       & 54.76 & 58.13 & 54.16 & 57.03 & 52.49 & 55.20 & 69.04 & 71.03 \\
Jina-v4 (Text) & 49.28 & 52.67 & 43.36 & 46.82 & 51.47 & 54.59 & 65.44 & 67.55 \\
Linq-Embed     & 60.72 & 63.45 & 49.22 & 52.50 & 49.82 & 53.01 & 69.70 & 71.39 \\
Jina-v4 (Vis)  & 54.47 & 58.40 & 57.62 & 60.77 & 71.99 & 74.47 & 80.36 & 81.84 \\
ColNomic    & \underline{60.85} & \underline{63.57} & 60.02 & 62.45 & 75.95 & 77.67 & 85.58 & 86.72 \\
ColEmbed & 60.12 & 63.16 & \underline{66.45} & \underline{68.92} & \textbf{77.07} & \textbf{79.08} & \underline{86.07} & \underline{87.08} \\
\midrule
\rowcolor{yellow!34}\textbf{\model} (ours) & \textbf{65.46} & \textbf{67.87} & \textbf{71.61} & \textbf{73.58} & \underline{76.49} & \underline{77.97} & \textbf{86.67} & \textbf{87.47} \\
\textit{\footnotesize (Std. error)} & \footnotesize$\pm$1.24 & \footnotesize$\pm$1.14 & \footnotesize$\pm$1.34 & \footnotesize$\pm$1.23 & \footnotesize$\pm$1.01 & \footnotesize$\pm$0.93 & \footnotesize$\pm$0.79 & \footnotesize$\pm$0.73 \\
\midrule
$\Delta\%$ vs. Best & \gain{$\uparrow$ 7.6\%} & \gain{$\uparrow$ 6.8\%} & \gain{$\uparrow$ 7.8\%} & \gain{$\uparrow$ 6.8\%} & \textcolor{red}{$\downarrow$ 0.8\%} & \textcolor{red}{$\downarrow$ 1.4\%} & \gain{$\uparrow$ 0.7\%} & \gain{$\uparrow$ 0.4\%} \\
\bottomrule
\end{tabular}
}
\caption{Fine-grained retrieval performance on the REAL-MM-RAG subsets.}
\label{tab:real_mm_rag_results}
\end{table}

\label{app:additional_results}

\section{Additional Results}
\section{Implementation Details. }

All experiments are run on an NVIDIA RTX 6000 GPU.

\subsection{Datasets}
\label{app:datasets}
We evaluate on four visual document retrieval benchmarks (Table~\ref{tab:dataset_stats}). In all of them, the retrieval unit is a rendered page, and each query must be answered by retrieving its relevant page(s) from the full corpus. 

\begin{table}[htbp]
\centering
\small
\resizebox{\columnwidth}{!}{%
\setlength{\tabcolsep}{6pt}
\begin{tabular}{llcrr}
\toprule
\textbf{Benchmark} & \textbf{Subset} & \textbf{Lang.} & \textbf{\#Docs} & \textbf{\#Pages} \\
\midrule
\multirow{8}{*}{ViDoRe~3}
 & Finance (EN)   & en & 6  & 2,942 \\
 & Comp.\ Science & en & 2  & 1,360 \\
 & Pharma         & en & 52 & 2,313 \\
 & HR             & en & 14 & 1,110 \\
 & Industrial     & en & 27 & 5,244 \\
 & Physics        & fr & 42 & 1,674 \\
 & Finance (FR)   & fr & 5  & 2,384 \\
 & Energy         & fr & 41 & 2,225 \\
\midrule
ViDoSeek & -- & en & 292 & 5,385 \\
SlideVQA & -- & en & 400 & 8,000 \\
\midrule
\multirow{4}{*}{REAL-MM-RAG}
 & FinReport  & en & 19 & 2,687 \\
 & FinSlides  & en & 65 & 2,280 \\
 & TechReport & en & 17 & 1,674 \\
 & TechSlides & en & 61 & 1,963 \\
\midrule
\textbf{Total} & & & \textbf{1,043} & \textbf{41,241} \\
\bottomrule
\end{tabular}
}
\caption{Statistics of the evaluation corpora. Language refers to the documents.}
\label{tab:dataset_stats}
\end{table}

\subsection{Document Retrieval Models}
Visual document retrieval methods fall into two families. \textbf{Text-mediated retrieval} first converts each page into a textual surrogate-via OCR, layout parsing, or VLM-generated descriptions-and then retrieves with a standard text encoder. This reuses mature text retrievers, but the conversion step discards spatial layout and visual content, and extraction errors propagate into the index. \textbf{Image-native multi-vector retrieval} removes the conversion step: a vision-language backbone encodes the page as patch-level token embeddings, the query is encoded as text tokens by the same model, and relevance is the late-interaction (MaxSim) sum introduced by ColBERT~\cite{khattab2020colbert} and brought to images by ColPali~\cite{ICLR2025_99e9e141}. This preserves layout and requires no OCR at query time, and has since been instantiated on several backbones. Its weakness is cost: representing each page with hundreds to thousands of token vectors makes exhaustive search expensive at scale. Table~\ref{tab:combined_model_methodology} lists the retrievers we evaluate across both families. 

\begin{table}[htbp]
\centering
\resizebox{\columnwidth}{!}{
\small
\setlength{\tabcolsep}{10pt}
\renewcommand{\arraystretch}{1.2}
\begin{tabular}{lrrr}
\toprule
% Using a nested tabular allows precise control over the vertical spacing
\textbf{Index Footprint} & \begin{tabular}[c]{@{}c@{}}\textbf{ViDoRe 3} \\[3pt] \footnotesize (19,252 pages)\end{tabular} 
                 & \begin{tabular}[c]{@{}c@{}}\textbf{ViDoSeek} \\[3pt] \footnotesize (5,385 pages)\end{tabular} 
                 & \begin{tabular}[c]{@{}c@{}}\textbf{SlideVQA} \\[3pt] \footnotesize (8,000 pages)\end{tabular} \\
\midrule
Nemotron-Embed-VL-1B         & 0.08 GB  & 0.02 GB  & 0.03 GB \\
ColEmbed-3B-v2   & 213.1 GB & 59.6 GB  & 88.6 GB \\
SCOUT Active        & 0.63 GB  & 0.58 GB  & 0.59 GB \\
\midrule
\textbf{Reduction}       & \textbf{337$\times$} & \textbf{104$\times$} & \textbf{151$\times$} \\
\bottomrule
\end{tabular}
}
\caption{Estimated active retrieval index footprint across corpora.}
\label{tab:index_footprint_corpora}
\end{table}

\begin{table}[h]
\centering
\resizebox{\columnwidth}{!}{
\begin{tabular}{lrrrr}
\toprule
\textbf{Parent retriever} & \textbf{Index (GB)} & \textbf{NDCG@10} & \textbf{Latency (ms)} \\
\midrule
Nemotron-Embed-VL-1B & 0.23 & 67.37 & 465.3 \\
ColNomic-7B & 6.35 & 67.70 & 777.7 \\
\bottomrule
\end{tabular}
}
\caption{Parent retriever ablation across 3 benchmarks: ViDoRe 3, SlideVQA, and ViDoSeek.}
\label{tab:first_stage_ablation}
\end{table}

\begin{table}[h]
\centering
\resizebox{\columnwidth}{!}{
\small
\setlength{\tabcolsep}{4pt}
\begin{tabular}{lrrrrrr}
\toprule
\textbf{Dataset} & \textbf{\#Docs} & \textbf{\#Pages} & \textbf{Time} & \textbf{Time/page} & \textbf{Cost/page} & \textbf{Total cost} \\
\midrule
ViDoRe 3     & 189 & 19,252 & 1h 40m & 312 ms & \$0.0023 & \$44.81 \\
SlideVQA     & 400 &  8,000 &    40m & 300 ms & \$0.0022 & \$17.78 \\
ViDoSeek     & 292 &  5,385 &    25m & 279 ms & \$0.0024 & \$12.78 \\
REAL-MM-RAG  & 162 &  8,604 &    45m & 314 ms & \$0.0025 & \$21.56 \\
\midrule
\textbf{Total} & \textbf{1,043} & \textbf{41,241} & \textbf{3h 30m} & -- & -- & \textbf{\$96.93} \\
\bottomrule
\end{tabular}
}
\caption{One-time offline cost of generating hierarchical page and document summaries with Gemini 2.5 Flash. Time is effective wall-clock time using 12 parallel API workers. Only newly added pages require summarization, so updates to a corpus scale with the number of changed pages rather than corpus size.}
\label{tab:offline_cost}
\end{table}

\begin{figure*}[h]
\centering
\setlength{\tabcolsep}{4px} 
\renewcommand{\arraystretch}{1}
\begin{tabular}{c p{0.43\linewidth} c p{0.43\linewidth}}
    \textbf{(a)} & \includegraphics[width=\linewidth,keepaspectratio]{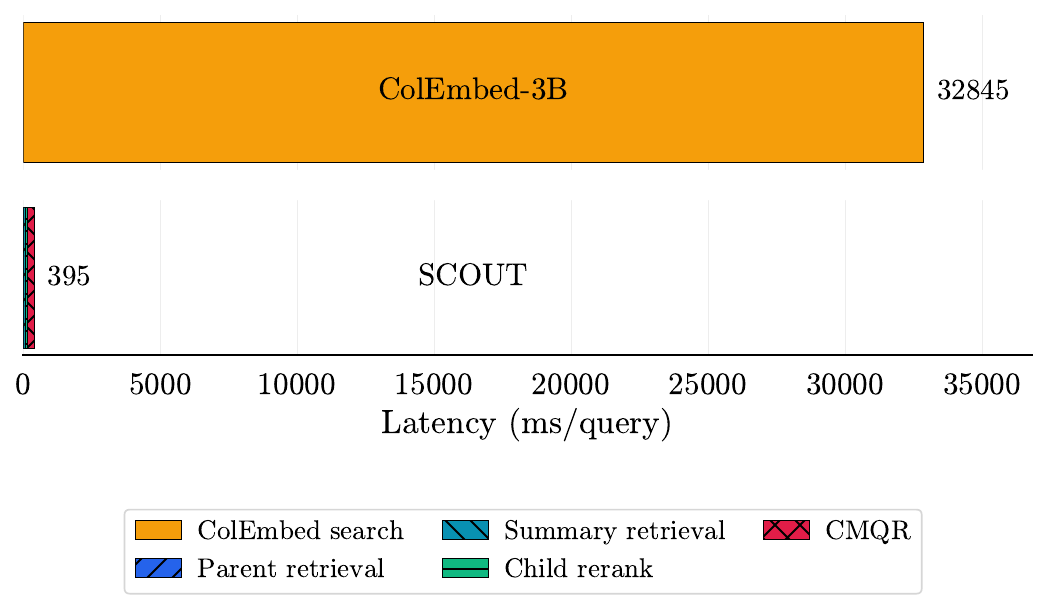} &
    \textbf{(b)} & \includegraphics[width=\linewidth,keepaspectratio]{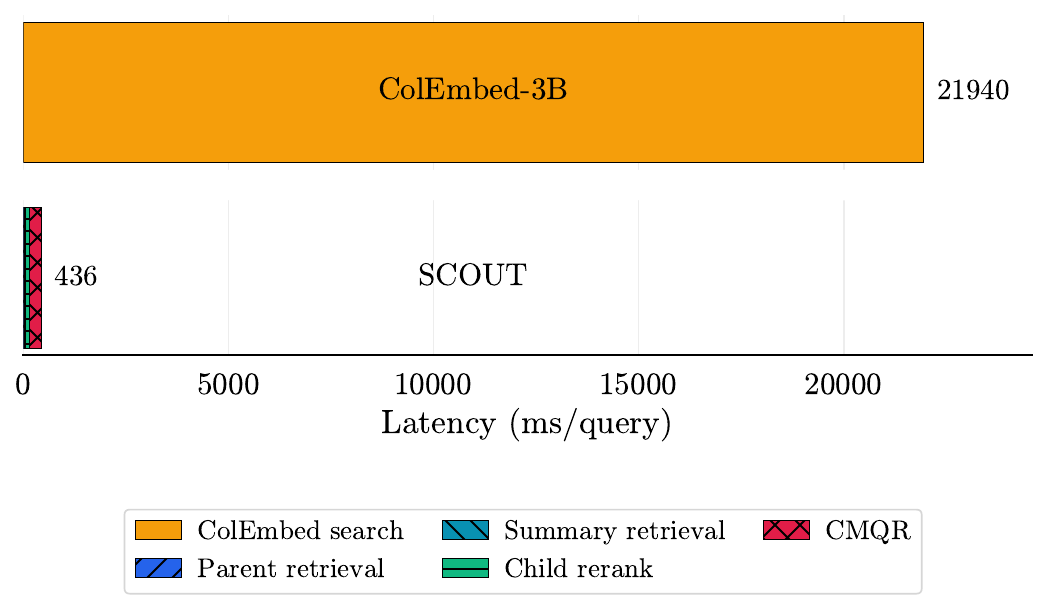} \\
\end{tabular}
\caption{Per-query latency breakdown comparing exhaustive ColEmbed-3B-v2 retrieval with our cascaded pipeline averaged over the \textbf{(a)} SlideVQA and \textbf{(b)} Vidoseek subsets.}
\label{fig:latency-plots-vidoseek-slidevqa}
\end{figure*}

\subsection{Offline Indexing}
\label{app:offline_indexing}
For each dataset and retrieval model, we build page-level indices before evaluation. 
The rendered page is the atomic retrieval unit: visual embeddings, OCR/page text, and summary embeddings are all keyed by the same page identifier. 
All page embeddings are pre-computed offline; reported latency therefore measures only query-time encoding, retrieval, reranking, fusion, and refinement.

For visual retrievers, we encode rendered page images directly. For text-only baselines, we index one OCR/page-text string per page. For ViDoRe v3, we use the benchmark-provided page-level Markdown text processed by the NeMo Retriever extraction service~\cite{nvidiaingest2024}. 
For ViDoSeek and SlideVQA we use page text extracted with MinerU\cite{wang2024mineru}. 

For the summary-based retrieval branch, we construct an additional structured textual view for each page. 
We generate page-level summaries using Gemini 2.5 Flash\cite{comanici2025gemini} and concatenate each page summary with the corresponding document-level description. 

Table~\ref{tab:index_footprint_corpora} reports the resulting index footprints. We distinguish total stored embeddings from the \emph{active} footprint-the representations that must be resident at query time.  ColEmbed-3B retrieval keeps the full multi-vector index active, scaling to hundreds of gigabytes even on these modestly sized corpora. In contrast, \model\ keeps only the lightweight first-stage index resident for corpus-wide search and accesses multi-vector embeddings solely for the bounded top-$K$ candidate set, reducing the active footprint.
 
\subsection{Offline Summarization Cost}
\label{app:offline_cost}
\model{} generates hierarchical page and document summaries once per corpus, at indexing time. Table~\ref{tab:offline_cost} reports this one-time cost for all evaluated corpora using Gemini 2.5 Flash. Summarization costs roughly \$0.0023 and 300\,ms of wall-clock time per page (with 12 parallel API workers). Indexing the full ViDoRe~3 corpus (19,256 pages) costs \$44.81 and takes about 1h 40m, and all four corpora (41,245 pages) together cost under \$100. The cost is incurred offline and never affects query-time latency, which makes it practical for enterprise settings where corpora are static or slowly updated. For frequently updated corpora, only newly added pages need to be summarized, so the update cost scales with the number of changed pages rather than with corpus size.

\begin{table}[t]
\centering
\resizebox{\columnwidth}{!}{
\begin{tabular}{llcccc}
\toprule
\textbf{Dataset} & \textbf{$\eta$} & \textbf{Steps} & $\alpha$ & \textbf{$T$} & \textbf{NDCG@10} \\ \midrule
\textit{ViDoRe (English)} & & & & & \\
\quad CS              & $1 \times 10^{-4}$ & 10 & 0.5 & 1.0 & 78.13 \\
\quad Finance         & $5 \times 10^{-5}$ & 50 & 0.5 & 1.0 & 66.86 \\
\quad Pharmaceuticals & $1 \times 10^{-3}$ & 25 & 0.5 & 1.0 & 66.99 \\
\quad HR              & $5 \times 10^{-5}$ & 25 & 0.5 & 1.0 & 64.12 \\
\quad Industrial      & $1 \times 10^{-4}$ & 10 & 0.5 & 1.0 & 51.09 \\ \midrule
\textit{ViDoRe (French)} & & & & & \\
\quad Finance         & $1 \times 10^{-4}$ & 25 & 0.5 & 1.0 & 47.58 \\
\quad Energy          & $5 \times 10^{-5}$ & 50 & 0.5 & 1.0 & 65.21 \\
\quad Physics         & $1 \times 10^{-4}$ & 25 & 0.5 & 1.0 & 50.46 \\ \midrule
\textit{REAL-MM-RAG} & & & & & \\
\quad FinSlides       & $5 \times 10^{-5}$ & 25 & 0.5 & 1.0 & 67.87 \\
\quad FinReport       & $5 \times 10^{-5}$ & 25 & 0.5 & 1.0 & 73.58 \\
\quad TechReport      & $1 \times 10^{-4}$ & 10 & 0.5 & 1.0 & 77.97 \\
\quad TechSlides      & $1 \times 10^{-5}$ & 25 & 0.5 & 1.0 & 87.47 \\ \midrule
\textit{Other Benchmarks} & & & & & \\
\quad ViDoSeek        & $1 \times 10^{-3}$ & 50 & 0.5 & 1.0 & 88.29 \\
\quad SlideVQA        & $1 \times 10^{-5}$ & 25 & 0.5 & 1.0 & 95.01 \\ \bottomrule
\end{tabular}
}
\caption{
Validation-selected CMQR hyperparameters across datasets at the main configuration ($K{=}25$, top-$T$ summary feedback $=20$). $\eta$ denotes the learning rate and $T$ the softmax temperature.}
\label{tab:optimal_hyperparams1}
\end{table}

\begin{table*}[t]
\centering
\begin{adjustbox}{max width=\linewidth} 
\begin{tabular}{l ccccc}
\toprule
\textbf{Dataset} & \textbf{\model} & \textbf{w/o Track A} & \textbf{w/o Track B} & \textbf{w/o Child} & \textbf{CMQR$\rightarrow$RRF} \\
\midrule
\multicolumn{6}{l}{\textit{ViDoRe 3}} \\
\midrule
\hspace{0.4em} Computer Science & \textbf{78.13} & 69.76 (\textcolor{red}{$\downarrow 12.00\%$}) & 75.65 (\textcolor{red}{$\downarrow 3.28\%$}) & 68.65 (\textcolor{red}{$\downarrow 13.81\%$}) & 74.66 (\textcolor{red}{$\downarrow 4.65\%$}) \\
\hspace{0.4em} Energy & \textbf{65.21} & 58.67 (\textcolor{red}{$\downarrow 11.15\%$}) & 60.97 (\textcolor{red}{$\downarrow 6.95\%$}) & 50.42 (\textcolor{red}{$\downarrow 29.33\%$}) & 61.91 (\textcolor{red}{$\downarrow 5.33\%$}) \\
\hspace{0.4em} Finance & \textbf{66.86} & 57.58 (\textcolor{red}{$\downarrow 16.12\%$}) & 62.46 (\textcolor{red}{$\downarrow 7.04\%$}) & 52.72 (\textcolor{red}{$\downarrow 26.82\%$}) & 63.09 (\textcolor{red}{$\downarrow 5.98\%$}) \\
\hspace{0.4em} Finance (fr) &\textbf{47.58} & 38.84 (\textcolor{red}{$\downarrow 22.50\%$}) & 39.80 (\textcolor{red}{$\downarrow 19.55\%$}) & 30.05 (\textcolor{red}{$\downarrow 58.34\%$}) & 40.54 (\textcolor{red}{$\downarrow 17.37\%$}) \\
\hspace{0.4em} HR & \textbf{64.12} & 54.03 (\textcolor{red}{$\downarrow 18.67\%$}) & 59.54 (\textcolor{red}{$\downarrow 7.69\%$}) & 49.49 (\textcolor{red}{$\downarrow 29.56\%$}) & 58.59 (\textcolor{red}{$\downarrow 9.44\%$}) \\
\hspace{0.4em} Industrial &\textbf{51.09} & 40.82 (\textcolor{red}{$\downarrow 25.16\%$}) & 46.05 (\textcolor{red}{$\downarrow 10.94\%$}) & 35.73 (\textcolor{red}{$\downarrow 42.99\%$}) & 45.27 (\textcolor{red}{$\downarrow 12.86\%$}) \\
\hspace{0.4em} Pharma & \textbf{66.99} & 63.44 (\textcolor{red}{$\downarrow 5.60\%$}) & 64.48 (\textcolor{red}{$\downarrow 3.89\%$}) & 58.28 (\textcolor{red}{$\downarrow 14.95\%$}) & 66.87 (\textcolor{red}{$\downarrow 0.18\%$}) \\
\hspace{0.4em} Physics & \textbf{50.46} & 46.66 (\textcolor{red}{$\downarrow 8.14\%$}) & 47.04 (\textcolor{red}{$\downarrow 7.27\%$}) & 42.61 (\textcolor{red}{$\downarrow 18.42\%$}) & 48.67 (\textcolor{red}{$\downarrow 3.68\%$}) \\
\midrule
\multicolumn{6}{l}{\textit{REAL-MM-RAG}} \\
\midrule
\hspace{0.4em} FinSlides & \textbf{67.87} & 54.03 (\textcolor{red}{$\downarrow 25.63\%$}) & 62.27 (\textcolor{red}{$\downarrow 8.99\%$}) & 48.02 (\textcolor{red}{$\downarrow 41.36\%$}) & 62.13 (\textcolor{red}{$\downarrow 9.24\%$}) \\
\hspace{0.4em} FinReport & \textbf{73.58} & 53.57 (\textcolor{red}{$\downarrow 37.37\%$}) & 66.83 (\textcolor{red}{$\downarrow 10.10\%$}) & 49.59 (\textcolor{red}{$\downarrow 48.38\%$}) & 64.27 (\textcolor{red}{$\downarrow 14.49\%$}) \\
\hspace{0.4em} TechReport & \textbf{77.97} & 59.84 (\textcolor{red}{$\downarrow 30.30\%$}) & 74.97 (\textcolor{red}{$\downarrow 4.00\%$}) & 58.95 (\textcolor{red}{$\downarrow 32.26\%$}) & 69.62 (\textcolor{red}{$\downarrow 11.99\%$}) \\
\hspace{0.4em} TechSlides & \textbf{87.47} & 73.41 (\textcolor{red}{$\downarrow 19.15\%$}) & 85.72 (\textcolor{red}{$\downarrow 2.05\%$}) & 76.11 (\textcolor{red}{$\downarrow 14.93\%$}) & 82.34 (\textcolor{red}{$\downarrow 6.23\%$}) \\
\midrule
ViDoSeek & \textbf{88.29} & 86.28 (\textcolor{red}{$\downarrow 2.33\%$}) & 85.60 (\textcolor{red}{$\downarrow 3.14\%$}) & 81.41 (\textcolor{red}{$\downarrow 8.45\%$}) & 88.19 (\textcolor{red}{$\downarrow 0.11\%$}) \\
SlideVQA & \textbf{95.01} & 91.76 (\textcolor{red}{$\downarrow 3.54\%$}) & 94.79 (\textcolor{red}{$\downarrow 0.23\%$}) & 92.79 (\textcolor{red}{$\downarrow 2.39\%$}) & 94.79 (\textcolor{red}{$\downarrow 0.23\%$}) \\
\midrule
Overall Average & \textbf{70.05} & 60.62 (\textcolor{red}{$\downarrow 15.55\%$}) & 66.16 (\textcolor{red}{$\downarrow 5.88\%$}) & 56.77 (\textcolor{red}{$\downarrow 23.38\%$}) & 65.78 (\textcolor{red}{$\downarrow 6.48\%$}) \\
\bottomrule
\end{tabular}
\end{adjustbox}
\caption{Component ablation across datasets (NDCG@10) comparing our full model to variants.}
\label{tab:component_ablation_appendix}
\end{table*}

\subsection{Hyperparameter Selection}
\label{app:hyperparameters}
We optimize only two hyperparameters for the Guided Query Refinement (GQR) module: the learning rate ($\eta$) and the number of optimization steps ($N_{\mathrm{steps}}$). We perform a grid search over the following parameter space:
\begin{gather*}
\eta \in \{1{\times}10^{-5}, 5{\times}10^{-5}, 1{\times}10^{-4}, \\
5{\times}10^{-4}, 1{\times}10^{-3}, 5{\times}10^{-3}\}, \\
N_{\mathrm{steps}} \in \{10, 25, 50\}.
\end{gather*}
As shown in Table~\ref{tab:optimal_hyperparams1} for each dataset, the optimal combination is selected based on NDCG@10 performance on a held-out 5\% validation split and subsequently fixed for the final evaluation. Following the average-consensus formulation of GQR \citep{uzan2026guided}, we fix the mixture weight to $\alpha = 0.5$, assigning equal importance to the primary visual distribution and the Qwen3 guidance distribution. We also fix the softmax temperature to $T = 1.0$. No hyperparameters are tuned on the final evaluation split.

\begin{figure}[t]
\centering
\includegraphics[width=\columnwidth]{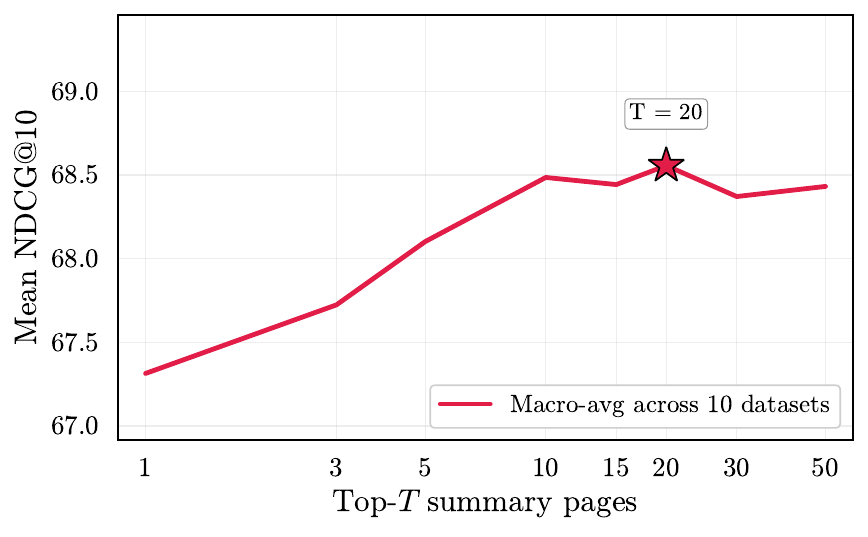}
\caption{Validation-set sweep over summary depth: mean NDCG@10 (macro-average over 3 datasets: ViDoRe 3, SlideVQA, and ViDoSeek.) for varying numbers of retrieved summary pages $T$.}
\label{fig:t_sweep}
\end{figure}

\paragraph{Summary depth $T$.}
We also select the number of summary pages $T$ that the semantic track contributes to the candidate pool on the validation split (Figure~\ref{fig:t_sweep}). Performance improves from $T{=}1$ to $T{=}10$ (about $+1.2$ NDCG@10) as more summary candidates enter the refinement pool, then plateaus: all settings with $T \geq 10$ stay within $0.2$ points of the peak at $T{=}20$. We therefore use $T{=}20$, which gives the best validation performance at negligible additional latency compared with the visual reranking and CMQR stages.

\paragraph{Optimizer and loss.}
GQR optimizes the multi-vector query embedding $z$ using Adam ($\beta_1=0.9$, $\beta_2=0.999$, $\epsilon=10^{-8}$).
The loss minimizes $\mathrm{KL}(p_{\mathrm{mix}} \,\|\, p_1(z^{(t)}))$, where
the consensus $p_{\mathrm{mix}} = \tfrac{1}{2}(p_1(z^{(0)}) + p_2)$ is computed
once at $t{=}0$ and held fixed (matching the reference implementation).
The initial $z^{(0)}$ is the ColEmbed-3B-v2 query embedding of the unmodified question.

\subsection{Domain and Language Generalization}
\label{app:domainLanguage}
Table \ref{tab:vidore_results} shows \texttt{ViDoRe-Finance}, \model\ achieves NDCG@5 of $64.27$ against $60.81$ for SOTA ($\Delta{=}+5.7$\%) at $22\times$ lower latency. The performance gain is consistent across the remaining ViDoRe subsets. On \texttt{ViDoRe-Physics}, NDCG@10 improves from $48.44$ to $50.46$ ($\Delta{=}+4.2$\%); on \texttt{ViDoRe-Finance (FR)}, NDCG@5 improves from $42.19$ to $44.42$ ($\Delta{=}+5.3$\%) as compared to the ColNomic; on \texttt{ViDoRe-HR}, NDCG@5 increases from $58.61$ to $61.49$ ($\Delta{=}+4.6$\%); and on \texttt{ViDoRe-Pharma},
NDCG@10 improves from $65.47$ to $66.99$ ($\Delta{=}+2.3$\%). Gains on \texttt{ViDoRe-Industrial}, \texttt{ViDoRe-CS}, and \texttt{ViDoRe-Energy} are more modest at $+2.9$\%, $+1.0$\%, and $+1.5$\% NDCG@5 points respectively, but remain positive throughout. \model's performance validates that the cascade architecture does not compromise retrieval quality for efficiency (c.f. Table \ref{tab:vidore_results}).

Table~\ref{tab:real_mm_rag_results} shows that \model{} achieves the largest gains on the Finance subsets of REAL-MM-RAG. On \texttt{FinReport}, NDCG@5 improves from $66.45$ to $71.61$ ($\Delta{=}+7.8$\%) over ColEmbed, and on \texttt{FinSlides} from $60.85$ to $65.46$ ($\Delta{=}+7.6$\%) over ColNomic, the strongest baseline on that subset. On the Technical subsets, where the baselines are already near ceiling, \model{} remains on par: \texttt{TechSlides} improves slightly ($+0.7$\% NDCG@5), while \texttt{TechReport} shows a small drop ($-0.8$\% NDCG@5) relative to ColEmbed. Text-only retrievers trail the visual models on every subset except FinSlides, where Linq-Embed is within $0.13$ points of the best visual baseline. Overall, \model{} matches or exceeds the strongest baseline on three of four subsets, with the largest gains on finance documents, whose pages are dense with tables and visually similar layouts.

\subsection{Downstream Performance}
\label{sec:downstream}
To assess whether retrieval gains carry over to downstream answer quality, we run an end-to-end QA evaluation with Qwen3-VL-8B-Instruct. According to Table~\ref{tab:vidore3_accuracy}), SCOUT improves over ColEmbed-3B-v2 on every subset, raising the average to 66.48 (+4.4\% relative), with the largest gains on Finance (+7.5\%) and French Finance (+8.1\%). 

\section{Practical Considerations}
\label{sec:pracical_consideration}
\model\ is designed to be modular and deployment-friendly. The {parent retriever} is interchangeable depending on enterprise requirements. Table \ref{tab:first_stage_ablation} shows how performance varies with different retrievers, at the cost of a 27$\times$ larger storage ($0.23$ GB vs. $6.35$ GB). For storage-constrained deployments, a smaller model (e.g. Nemotron-Embed-VL-1B) is a practical choice, however even replacing it with larger models (e.g. ColNomic-7B), the performance sees only a bare minimum gain. Likewise, the {\em child reranker} can be replaced by any enterprise-specific reranker, with the only constraint that it produces embeddings to support the downstream similarity function in CMQR. This storage-latency trade-off gives a clear axis for tuning \model\ with deployment constraints.

\section{Reciprocal Rank Fusion}
\label{app:rrf_details}
We use Reciprocal Rank Fusion (RRF) as a non-parametric rank-level fusion baseline between the visual retrieval branch and the summary-based text retrieval branch. Given two ranked lists of pages, $\pi^{\mathrm{vis}}_q$ and $\pi^{\mathrm{sum}}_q$, RRF assigns each candidate page $p$ the score

\begin{table}[t]
\centering
\resizebox{\columnwidth}{!}{
\begin{tabular}{lllr}
\toprule
\textbf{Split} & \textbf{Dataset} & \textbf{Method} & \textbf{NDCG@10} \\
\midrule
\multirow{9}{*}{Single-hop} 
 & \multirow{3}{*}{SlideVQA} 
 & Parent retrieval & 93.87 \\
 & & +Child reranker & 95.44 \\
 & & \textbf{\model} & \textbf{95.59} \\
\cmidrule{2-4}
 & \multirow{3}{*}{ViDoRe 3} 
 & Parent retrieval & 49.37 \\
 & & + Child reranker & 58.05 \\
 & & \textbf{\model} & \textbf{62.41} \\
\cmidrule{2-4}
 & \multirow{3}{*}{ViDoSeek} 
 & Parent retrieval & 76.82 \\
 & & + Child reranker & 81.33 \\
 & & \textbf{\model} & \textbf{86.11} \\
\midrule
\multirow{9}{*}{Multi-hop} 
 & \multirow{3}{*}{SlideVQA} 
 & Parent retrieval & 89.72 \\
 & & + Child reranker & 92.92 \\
 & & \textbf{\model} & \textbf{93.37} \\
\cmidrule{2-4}
 & \multirow{3}{*}{ViDoRe 3} 
 & Parent retrieval & 39.69 \\
 & & + Child reranker & 46.03 \\
 & & \textbf{\model} & \textbf{49.67} \\
\cmidrule{2-4}
 & \multirow{3}{*}{ViDoSeek} 
 & Parent retrieval  & 87.46 \\
 & & + Child reranker & 91.22 \\
 & & \textbf{\model} & \textbf{91.17} \\
\bottomrule
\end{tabular}
}
\caption{Single-hop and multi-hop performance analysis (NDCG@10) across SlideVQA, ViDoRe, and ViDoSeek. ViDoRe values are macro-averaged across its 8 sub-datasets.}
\label{tab:single_multi_analysis_all}
\end{table}

\begin{equation}
    s_{\mathrm{RRF}}(q,p)
    =
    \sum_{\pi \in \{\pi^{\mathrm{vis}}_q, \pi^{\mathrm{sum}}_q\}}
    \frac{1}{k + \operatorname{rank}_{\pi}(p)},
\end{equation}
where $\operatorname{rank}_{\pi}(p)$ is the 1-indexed rank of page $p$ in list $\pi$, and $k$ is a smoothing constant. We set $k=60$, following the standard RRF setting. RRF is applied to the union of pages appearing in either list, and pages are sorted by $s_{\mathrm{RRF}}(q,p)$ to obtain the fused ranking. The report analysis compared to CMQR shown in Table\ref{tab:component_ablation_appendix}. 

\begin{table*}[t]
\centering
\setlength{\tabcolsep}{3.4pt}
\scalebox{0.8}{%
\begin{tabular}{lccccccccc}
\toprule
\textbf{Model} & \textbf{CS} & \textbf{Finance} & \textbf{Pharma} & \textbf{HR} & \textbf{Ind.} & \textbf{Fin. (FR)} & \textbf{Energy} & \textbf{Physics} & \textbf{Avg.} \\
\midrule
ColEmbed-3B-v2 & 79.76 & 58.69 & 65.46 & 60.25 & 58.97 & 52.08 & 64.73 & 69.23 & 63.65 \\
\rowcolor{yellow!34} \model\ + page summaries & \textbf{82.52} & \textbf{63.09} & \textbf{68.01} & \textbf{63.74} & \textbf{59.85} & \textbf{56.28} & \textbf{67.83} & \textbf{70.49} & \textbf{66.48} \\
\midrule
$\Delta\%$ vs. ColEmbed & \gain{$\uparrow$ 3.5\%} & \gain{$\uparrow$ 7.5\%} & \gain{$\uparrow$ 3.9\%} & \gain{$\uparrow$ 5.8\%} & \gain{$\uparrow$ 1.5\%} & \gain{$\uparrow$ 8.1\%} & \gain{$\uparrow$ 4.8\%} & \gain{$\uparrow$ 1.8\%} & \gain{$\uparrow$ 4.4\%} \\
\bottomrule
\end{tabular}
}
\caption{End-to-end answer quality on the ViDoRe~3 subsets, measured as the average judge score (Correct $=1$, Partially correct $=0.5$, Incorrect $=0$). Generator: Qwen3-VL-8B-Instruct, top-5 retrieved pages; judge: GPT-5.2.}
\label{tab:vidore3_accuracy}
\end{table*}

\subsection{Single-hop vs. Multi-hop Complexity}
\label{sec:analysis:single_multi}

To examine how \model\ behaves under different query complexities, we evaluate three stages of the pipeline---parent retrieval, the child reranker, and the full \model---on single-hop and multi-hop query subsets separately (Table~\ref{tab:single_multi_analysis_all}).

\begin{table*}[ht]
\centering
\scalebox{0.8}{
\begin{tabular}{l|l|c|c|c}
\toprule
\textbf{Benchmark} & \textbf{Subset} & \textbf{Method} & \textbf{NDCG@10} ($\uparrow$) & \textbf{Latency (ms)} ($\downarrow$) \\
\midrule
\multirow{3}{*}{SlideVQA} & \multirow{3}{*}{--} & \model\ & 95.01 & 406.89 \\
 & & ColNomic-7B & 94.57 & 331.01 \\
 & & ColEmbed-3B-v2 & 94.86 & 32,844.94 \\
\midrule
\multirow{24}{*}{ViDoRe} & \multirow{3}{*}{cs} & \model\ & 78.13 & 387.43 \\
 & & ColNomic-7B & 76.93 & 98.89 \\
 & & ColEmbed-3B-v2 & 77.22 & 7,088.78 \\
\cmidrule{2-5}
 & \multirow{3}{*}{energy} & \model\ & 65.21 & 553.34 \\
 & & ColNomic-7B & 63.69 & 108.45 \\
 & & ColEmbed-3B-v2 & 64.17 & 10,234.83 \\
\cmidrule{2-5}
 & \multirow{3}{*}{finance} & \model\ & 66.86 & 530.34 \\
 & & ColNomic-7B & 56.88 & 134.24 \\
 & & ColEmbed-3B-v2 & 63.41 & 12,149.66 \\
\cmidrule{2-5}
 & \multirow{3}{*}{finance\_fr} & \model\ & 47.58 & 456.57 \\
 & & ColNomic-7B & 45.89 & 125.14 \\
 & & ColEmbed-3B-v2 & 43.38 & 12,047.96 \\
\cmidrule{2-5}
 & \multirow{3}{*}{hr} & \model\ & 64.12 & 539.79 \\
 & & ColNomic-7B & 59.91 & 72.35 \\
 & & ColEmbed-3B-v2 & 61.32 & 4,554.36 \\
\cmidrule{2-5}
 & \multirow{3}{*}{industrial} & \model\ & 51.09 & 454.68 \\
 & & ColNomic-7B & 49.56 & 207.64 \\
 & & ColEmbed-3B-v2 & 49.85 & 21,511.19 \\
\cmidrule{2-5}
 & \multirow{3}{*}{pharma} & \model\ & 66.99 & 448.69 \\
 & & ColNomic-7B & 62.89 & 111.86 \\
 & & ColEmbed-3B-v2 & 65.47 & 11,757.60 \\
\cmidrule{2-5}
 & \multirow{3}{*}{physics} & \model\ & 50.46 & 422.87 \\
 & & ColNomic-7B & 48.44 & 93.00 \\
 & & ColEmbed-3B-v2 & 47.18 & 8,735.13 \\
\midrule
\multirow{3}{*}{ViDoSeek} & \multirow{3}{*}{--} & \model\ & 88.29 & 452.64 \\
 & & ColNomic-7B & 85.03 & 272.17 \\
 & & ColEmbed-3B-v2 & 85.76 & 21,939.95 \\
\midrule
\multirow{12}{*}{REAL-MM-RAG} & \multirow{3}{*}{FinSlides} & \model\ & 67.87 & 466.64 \\
 & & ColNomic-7B & 63.57 & 109.90 \\
 & & ColEmbed-3B-v2 & 63.16 & 10,965.30 \\
\cmidrule{2-5}
 & \multirow{3}{*}{FinReport} & \model\ & 73.58 & 531.46 \\
 & & ColNomic-7B & 62.45 & 196.00 \\
 & & ColEmbed-3B-v2 & 68.92 & 15,199.30 \\
\cmidrule{2-5}
 & \multirow{3}{*}{TechReport} & \model\ & 77.97 & 452.66 \\
 & & ColNomic-7B & 77.67 & 106.40 \\
 & & ColEmbed-3B-v2 & 79.08 & 9,025.10 \\
\cmidrule{2-5}
 & \multirow{3}{*}{TechSlides} & \model\ & 87.47 & 450.60 \\
 & & ColNomic-7B & 86.72 & 152.80 \\
 & & ColEmbed-3B-v2 & 87.08 & 10,447.00 \\
\midrule
\midrule
\multirow{3}{*}{Average} & \multirow{3}{*}{--} & \textbf{\model} & \textbf{70.05} & {468.19} \\
 & & {ColNomic-7B} & {66.73} & {151.42} \\
 & & {ColEmbed-3B-v2} & {67.92} & {13,464.36} \\
\midrule
\multirow{3}{*}{{Speedup vs. ColEmbed}} & \multirow{3}{*}{--} & \textbf{\model} & -- & \textbf{28.76$\times$} \\
 & & {ColNomic-7B} & -- & \textbf{88.92$\times$} \\
 & & {ColEmbed-3B-v2} & -- & {1.00$\times$} \\
\bottomrule
\end{tabular}
}
\caption{Retrieval performance, latency, and relative speedups grouped by dataset and method.}
\label{tab:latency_detailed}
\end{table*}

Across both query types, the child reranker consistently improves over parent retrieval, confirming the value of bounded late-interaction reranking. 
On ViDoRe single-hop queries, NDCG@10 increases from $49.37$ to $58.05$ after reranking ($\Delta{=}+8.68$), and the full pipeline further improves to $62.41$ ($\Delta{=}+4.36$ over the reranker). The same trend holds on ViDoRe multi-hop queries, where performance rises from $39.69$ to $46.03$ and then to $49.67$. 
SlideVQA shows a similar pattern, with the full pipeline achieving the best score on both single-hop and multi-hop subsets. 
On ViDoSeek, \model\ improves strongly over parent retrieval on both splits, although on the multi-hop subset the full pipeline is effectively tied with the child reranker ($91.17$ vs. $91.22$).

Multi-hop queries are harder on ViDoRe and SlideVQA, where full-pipeline NDCG@10 drops from $62.41$ to $49.67$ and from $95.59$ to $93.37$, respectively. 
ViDoSeek is an exception: its multi-hop subset scores higher than its single-hop subset across all stages, suggesting that query difficulty is dataset-dependent rather than determined only by hop count. 
Overall, the cascade remains effective across both regimes, with the largest gains appearing on the more challenging ViDoRe 3 split.

\begin{figure}[t]
\centering
\begin{subfigure}{\columnwidth}
  \includegraphics[width=\linewidth]{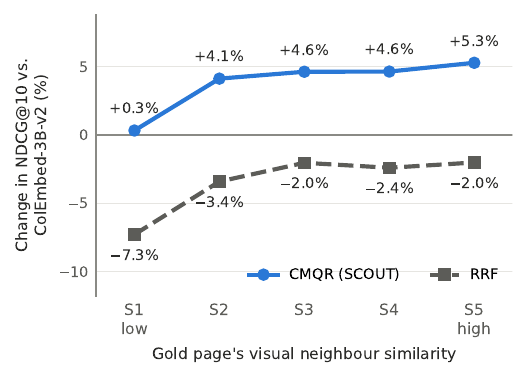}
  \caption{By the gold page's visual neighbour similarity.}
  \label{fig:gain_similarity}
\end{subfigure}
\begin{subfigure}{\columnwidth}
  \includegraphics[width=\linewidth]{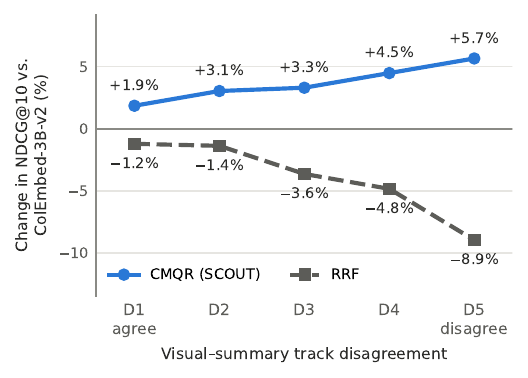}
  \caption{By visual--summary track disagreement.}
  \label{fig:gain_disagreement}
\end{subfigure}
\caption{Relative NDCG@10 change over full-corpus ColEmbed-3B-v2 for CMQR (\model{}) and RRF on the same candidate union, by quintile (computed within each of the 12 subsets; 17,162 queries).}
\label{fig:cmqr_rrf_quintiles}
\end{figure}

\paragraph{Robustness experiment setup.}
\label{app:robustness_setup}
We test how retrieval changes when the same query is reworded, keeping gold pages, corpus and hyperparameters identical to the main experiments. For REAL-MM-RAG, we use the released level-3 rephrasings of all 4{,}099 evaluation queries. For ViDoRe~3, ViDoSeek and SlideVQA, which release no rephrasings, we select the queries whose wording overlaps most with the gold page's text, excluding names, numbers and titles (100 queries per ViDoRe~3 subset, 300 each for ViDoSeek and SlideVQA). This text-only criterion is fixed before rephrasing. GPT-5.2 rephrases each query with REAL-MM-RAG's level-3 prompt. We keep a rephrasing only if it preserves all numbers and identifiers and a GPT-5.2 as a verifier judges it equivalent, non-leaking and meaningfully reworded (Figure~\ref{fig:paraphrase_prompt}); we resample a failed rephrasing up to three times. This leaves 792, 296 and 293 queries, whose rephrasings remove 22-34\% of the query words that match the gold page (35\% for REAL-MM-RAG). Each method is run on both the original and the rephrased version of every query, and we test whether the drop in NDCG@10 differs between methods.

\section{Generative AI Use}
We used generative AI assistants for grammatical revision, stylistic polishing, and as a coding assistant. Our use of generative AI did not amount to any substantive outsourcing of the intellectual work.

\onecolumn
\clearpage
\section{Document-Level Summary Prompt}
\label{appendix:doc_indexing_prompt}

\begin{figure*}[ht!]
\begin{tcolorbox}[
    colback=blue!5!white,
    colframe=blue!50!black,
    title=Document-level indexing prompt,
    fontupper=\footnotesize
]

You are a Document Indexing Expert. Your goal is to convert a document into a compact semantic descriptor that helps disambiguate it from other similar documents in a multi-document retrieval setting.

\vspace{1em}
\textbf{Instructions:}
\begin{enumerate}[leftmargin=*,itemsep=2pt]
    \item Analyze ALL pages of the provided document (provided as images in order).
    \item Extract the exact document title, authors, publication venue, and year from the cover/header --- these are critical for disambiguation.
    \item \texttt{key\_entities} must contain UNIQUE PROPER NOUNS that appear only in this document --- specific model names, dataset names, company names, author names, algorithm names. Avoid generic terms like \textit{``Large Language Models''} or \textit{``deep learning''}.
    \item Respond ONLY with a valid JSON object --- no explanation, no markdown, no code block.
\end{enumerate}

\vspace{1em}
\textbf{Schema:}

\begin{Verbatim}[breaklines=true, frame=single, fontsize=\scriptsize]
{
  "doc_id":  "exact title as it appears in the document",
  "authors": "author names if present, else null",
  "venue":   "journal, conference, or publication name if present,
              else null",
  "year":    "publication year if present, else null",
  "global_summary": "2-3 sentences: document type, exact topic, unique
                     contribution or finding. Be specific - mention the
                     actual method name, dataset name, or company name.",
  "key_entities": ["8-12 unique proper nouns that identify THIS document
                    - specific model names, dataset names, company names,
                    author surnames, algorithm names, metric names"]
}
\end{Verbatim}

\vspace{0.5em}
\end{tcolorbox}
\caption{Prompt used to generate document-level semantic summaries. At indexing time, the JSON fields shown are concatenated and prepended to each page summary before embedding; \texttt{global\_summary} is truncated to 300 characters and \texttt{key\_entities} to the first 15 entries.}
\label{fig:doc_indexing_prompt}
\end{figure*}

\begin{table*}[t]
\centering
\small
\resizebox{\textwidth}{!}{
\begin{tabular}{lllrrrl}
\toprule
\textbf{Model alias} &
\textbf{Full model name} &
\textbf{Size} &
\textbf{Vecs/Pg} &
\textbf{Dim} &
\textbf{Footprint} &
\textbf{HF Identifier} \\
& & (Params) & & & (GiB/1M Pgs) & \\
\midrule

Linq-Embed &
Linq-Embed-Mistral &
7.1B & 1 & 4096 & 7.6 &
\href{https://huggingface.co/Linq-AI-Research/Linq-Embed-Mistral}{\nolinkurl{Linq-AI-Research/Linq-Embed-Mistral}} \\

Qwen3-8B &
Qwen3-Embedding-8B &
8B & 1 & 4096 & 7.6 &
\href{https://huggingface.co/Qwen/Qwen3-Embedding-8B}{\nolinkurl{Qwen/Qwen3-Embedding-8B}} \\

Jina-v4 (Text) &
jina-embeddings-v4 &
3.8B & 1 & 2048 & 3.8 &
\href{https://huggingface.co/jinaai/jina-embeddings-v4}{\nolinkurl{jinaai/jina-embeddings-v4}} \\

Jina-v4 (Vision) &
jina-embeddings-v4 &
3.8B & $\sim$755 & 128 & 180.0 &
\href{https://huggingface.co/jinaai/jina-embeddings-v4}{\nolinkurl{jinaai/jina-embeddings-v4}} \\

ColNomic-7B &
ColNomic-Embed-Multimodal-7B &
7B & $\sim$755 & 128 & 180.0 &
\href{https://huggingface.co/nomic-ai/colnomic-embed-multimodal-7b}{\nolinkurl{nomic-ai/colnomic-embed-multimodal-7b}} \\

ColEmbed-3B-v2 &
Llama-Nemotron-ColEmbed-VL-3B-v2 &
3B & 1,802 & 3072 & 10,311.1 &
\href{https://huggingface.co/nvidia/llama-nemotron-colembed-vl-3b-v2}{\nolinkurl{nvidia/llama-nemotron-colembed-vl-3b-v2}} \\

Nemotron-Embed-VL-1B &
Llama-Nemotron-Embed-VL-1B-v2 &
1B & 1 & 2048 & 3.8 &
\href{https://huggingface.co/nvidia/llama-nemotron-embed-vl-1b-v2}{\nolinkurl{nvidia/llama-nemotron-embed-vl-1b-v2}} \\

\bottomrule
\end{tabular}
}
\caption{Specification of the evaluated retrieval models. We list model aliases, full model names, parameter counts, page-level embedding structure, estimated index footprint, and Hugging Face repositories. Footprint is estimated for one million pages using 16-bit precision and excludes metadata, file-system, and indexing overhead.}
\label{tab:combined_model_methodology}
\end{table*}

\clearpage
\section{Page-Level Summary Prompt}
\label{appendix:page_indexing_prompt}

\begin{figure*}[ht!]
\titleformat{\subsection}{\bfseries\large}{D.\arabic{section}}{1em}{}
\begin{tcolorbox}[colback=orange!5!white, colframe=orange!50!black, title=Page-level indexing prompt]
You are an expert document indexer. Your task is to analyze this document page and write a dense, highly searchable description optimized for a vector-based embedding search engine.

\vspace{1em}
\textbf{Rules:}
\begin{itemize}[leftmargin=*,itemsep=2pt]
    \item Prevent token bloat: If a table is massive, do not list every cell. Capture the schema and the main takeaway.
    \item Output ONLY the structured text below. Do not include any introductory or concluding conversational filler.
\end{itemize}

\vspace{1em}
\textbf{Structure your output exactly as follows:}

\vspace{1em}
\textbf{PAGE OVERVIEW:}\\
Write a 2--3 sentence summary of the page's core semantic intent. State clearly what topics are covered, what specific claims or conclusions are made, and what hypothetical questions this page perfectly answers.

\vspace{1em}
\textbf{STRUCTURAL CONTENT \& KEY DATA:}\\
For tables, charts, figures, or code, extract the schema and the narrative takeaway, NOT the raw data grid:
\begin{itemize}[leftmargin=*,itemsep=2pt]
    \item \textbf{Tables:} State the table title, column/row categories, and describe the notable trends. State exact numbers ONLY for major totals, minimums/maximums, or critical metrics.
    \item \textbf{Charts/Graphs:} Describe the chart type, axes, and the visual trend or pattern shown.
    \item \textbf{Figures/Diagrams:} Describe the core process or architecture shown. List the main components and use action verbs to describe how they relate or flow into one another.
    \item \textbf{Formulas/Code:} State exactly what the formula calculates or the code executes.
\end{itemize}
\textit{(If there are no structural elements, write ``No tables, charts, figures, or formulas present.'')}

\vspace{1em}
\textbf{NAMED ENTITIES \& IDENTIFIERS:}\\
Provide a concise, comma-separated list of specific, highly searchable keywords found on the page, grouped by:
\begin{itemize}[leftmargin=*,itemsep=2pt]
    \item Organizations \& People:
    \item Locations:
    \item Products \& Technologies:
    \item Dates \& Key Metrics: \textit{(include exact rates, percentages, dollar amounts)}
    \item Identifiers (Acronyms, Tickers, Product Codes):
\end{itemize}

\end{tcolorbox}
\caption{Prompt used to generate page-level semantic summaries for vector-based retrieval. Each page is summarized into a structured text representation comprising a high-level overview, schema-level descriptions of tables/charts/figures, and grouped named entities. }
\label{fig:page_indexing_prompt}
\end{figure*}

\section{Query Paraphrase Prompts}
\label{appendix:paraphrase_prompt}

\begin{minipage}{\linewidth}
\begin{tcolorbox}[colback=gray!5!white, colframe=gray!60!black, title={Rephrasing prompt (REAL-MM-RAG level 3, verbatim)}]
You are an expert at rephrasing questions while retaining their meaning.
Your task is to significantly rewrite the following question using different words and sentence order, maintaining the same intent and focus while changing a lot:

\vspace{0.5em}
Question: ``\{query\}''

\vspace{0.5em}
Write only the Rephrased Question.
\end{tcolorbox}

\vspace{0.5em}

\begin{tcolorbox}[colback=blue!5!white, colframe=blue!50!black, title=Paraphrase verification prompt]
You verify query pairs for an information-retrieval robustness benchmark.

\vspace{1em}
Determine whether the candidate is a meaningfully rewritten but strictly semantically equivalent version of the original. It must preserve the expected answer type, entities, numbers, dates, units, technical terms, comparisons, negations, alternatives, conjunctions, and multi-step requirements. It must not add contextual clues, resolve implicit references, disclose an answer, broaden the query, or narrow the query.

\vspace{1em}
Return valid JSON only:\\
\texttt{\{"equivalent": true, "details\_preserved": true,\\ "answer\_leakage": false, "meaningfully\_rewritten": true,\\ "reason": "brief explanation"\}}
\end{tcolorbox}
\captionof{figure}{Prompts used to construct the rephrasing-robustness evaluation sets for ViDoRe~3, ViDoSeek and SlideVQA. Each original query is rewritten with REAL-MM-RAG's level-3 rephrasing prompt, used verbatim (top). A candidate is kept only if it preserves every number and identifier of the original and passes an independent verification call (bottom); otherwise it is regenerated, up to three attempts, and queries without an accepted rephrasing are discarded. Both prompts use GPT-5.2. REAL-MM-RAG itself is evaluated with the benchmark's released level-3 rephrasings.}
\label{fig:paraphrase_prompt}
\end{minipage}

\end{document}